\documentclass[10pt,twocolumn,letterpaper]{article}

\usepackage{cvpr}
\usepackage{times}
\usepackage{epsfig}
\usepackage{graphicx}
\usepackage{amsmath}
\usepackage{amssymb}

\usepackage{color}
\usepackage{mathtools}

\usepackage{verbatim}

\usepackage{setspace}
\usepackage{algorithm}
\usepackage[noend]{algpseudocode}
\usepackage{enumitem}
\usepackage{xspace}
\usepackage{xparse}
\usepackage{csvsimple}

\newcommand{\ud}{\,\mathrm{d}}

\newcommand{\R}{\mathbb{R}}

\newcommand{\dv}[1]{\mbox{div}\left( {#1} \right)}

\newcommand{\argmin}{\arg\!\min}
\newcommand{\argmax}{\arg\!\max}
\newcommand{\prox}{\textrm{prox}}

\DeclareDocumentCommand\newstep{o}{%
\item\IfNoValueTF{#1}{}{#1 \textendash\xspace}}

\newlist{steps}{enumerate}{1}
\setlist[steps]{label=\textit{Step \arabic*:},leftmargin=*}

\graphicspath{{./figure/}}

\usepackage[pagebackref=true,breaklinks=true,letterpaper=true,colorlinks,bookmarks=false]{hyperref}

\cvprfinalcopy 

\ifcvprfinal\fi
\begin{document}

\title{Multi-Label Segmentation via Residual-Driven Adaptive Regularization}

\author{Byung-Woo Hong\qquad Ja-Keoung Koo\\
Chung-Ang University, Korea\\
{\tt\small \{hong,jakeoung\}@cau.ac.kr}
\and
Stefano Soatto\\
University of California Los Angeles, U.S.A.\\
{\tt\small soatto@cs.ucla.edu}
}

\maketitle

\begin{abstract}
We present a variational multi-label segmentation algorithm based on a robust Huber loss for both the data and the regularizer, minimized within a convex optimization framework. We introduce a novel constraint on the common areas, to bias the solution towards mutually exclusive regions.
We also propose a regularization scheme that is adapted to the spatial statistics of the residual at each iteration, resulting in a varying degree of regularization being applied as the algorithm proceeds: the effect of the regularizer is strongest at initialization, and wanes as the solution increasingly fits the data. This minimizes the bias induced by the regularizer at convergence. We design an efficient convex optimization algorithm based on the alternating direction method of multipliers using the equivalent relation between the Huber function and the proximal operator of the one-norm. We empirically validate our proposed algorithm on synthetic and real images and offer an information-theoretic derivation of the cost-function that highlights the modeling choices made. 
\end{abstract}

\section{Introduction}

To paraphrase the statistician Box, there is no such thing as a {\em wrong} segmentation. Yet, partitioning the image domain into multiple regions that exhibit some kind of homogeneity is {\em useful} in a number of subsequent stages of visual processing. So much so that segmentation remains an active area of research, with its own benchmark datasets that measure how {\em right} a segmentation is, often in terms of congruence with human annotators (who themselves are often incongruent).

The method of choice is to select a model, or cost function, that tautologically {\em defines} what a right segmentation is, and then find it via optimization. Thus, most segmentation methods are optimal, just with respect to different criteria.
Classically, one selects a model by picking a function(al) that measures {\em data fidelity}, which can be interpreted probabilistically as a log-likelihood, and one that measures {\em regularity}, which can be interpreted as a prior, with a parameter that trades off the two. In addition, since the number of regions is not only unknown, but also undefined (there could be any number of regions between one and the number of pixels in a single image), typically there is a {\em complexity cost} that drives the selection of a model among many.
Specifically for the case of multi-label, multiply-connected region-based segmentation, there is a long and illustrious history of contributions too long to list here, but traceable to~\cite{vese2002multiphase,grady2004multi,micusik2007multi,pock2008convex,nieuwenhuis2013survey} and references therein.
%
%
\def\fh{77pt}
\begin{figure}
\centering
\begin{tabular}{c@{}c}
\includegraphics[totalheight=\fh]{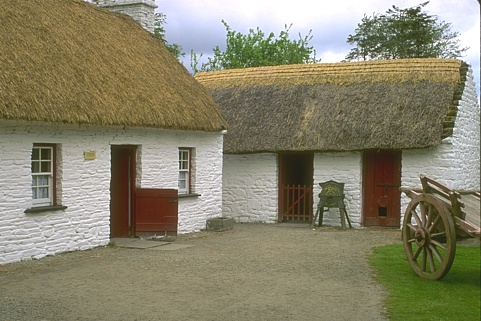} &
\includegraphics[totalheight=\fh]{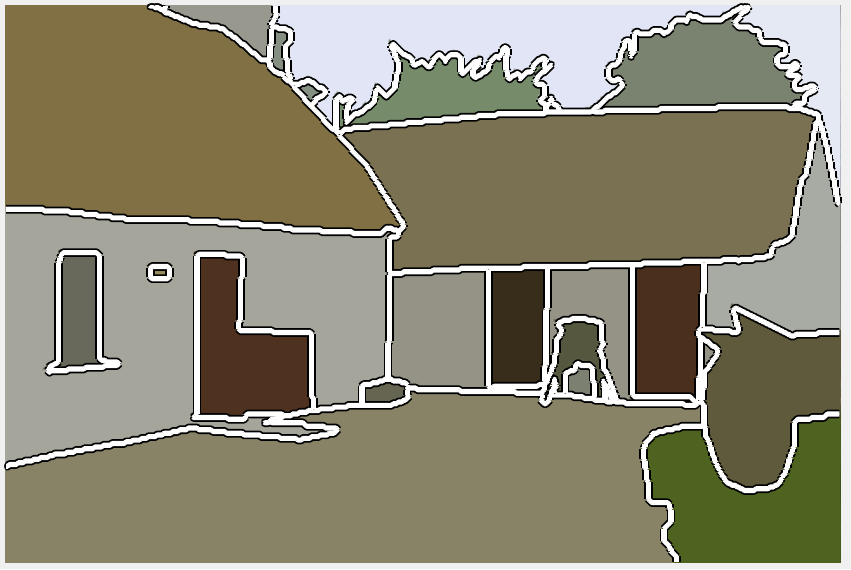}\\
(a) input image &
(b) segmentation\\
\vspace{-7pt}
\\
\includegraphics[totalheight=\fh]{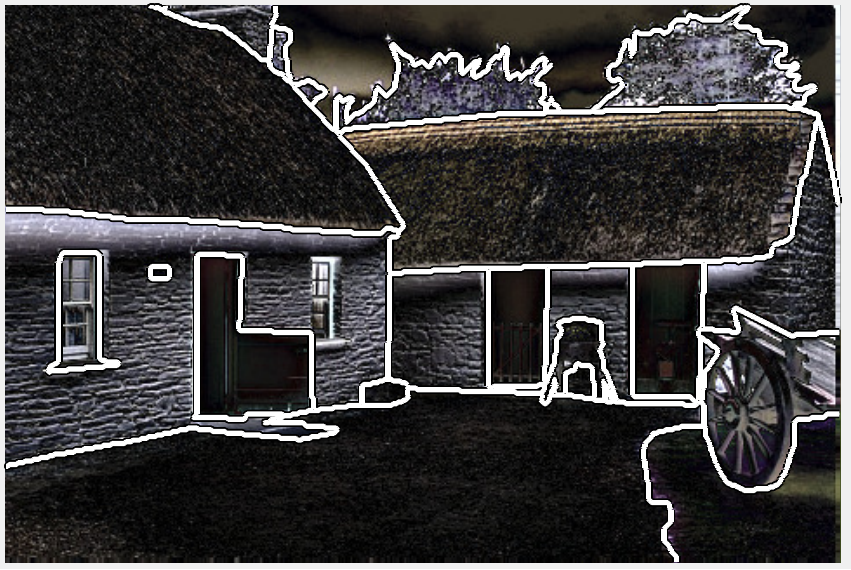} &
\includegraphics[totalheight=\fh]{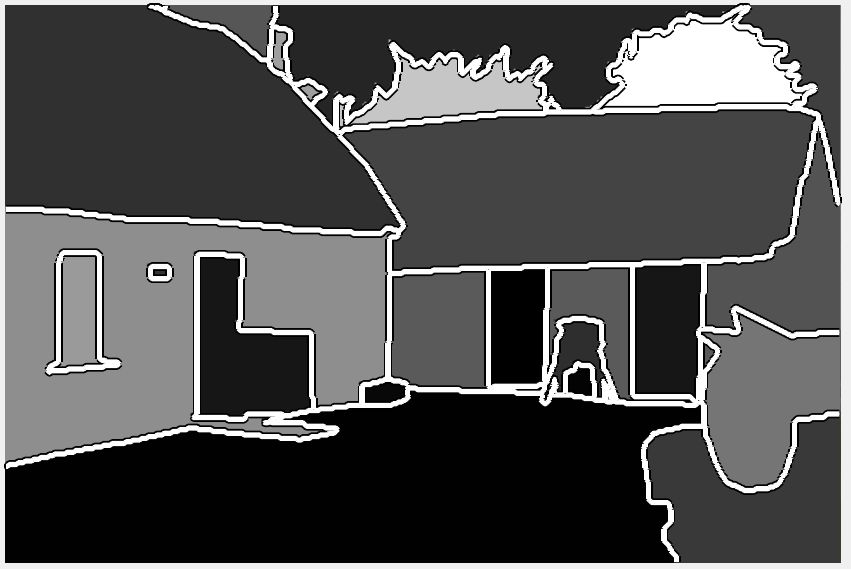}\\
(c) residual & 
(d) residual variance
\end{tabular}
\caption{Sample residual (c) and variance (d) for a segmentation (b) of an image (a) are not constant: Bright means large variance (d) or residual (c), and dark means small. Yet, most multi-label segmentation methods assume constant statistics and residual variance.}
\label{fig:motivation}
\end{figure}
%
%

In each and every one of these works, to the best of our knowledge, the trade-off between data fidelity and regularization is determined from the distribution of the optimization residual and assumed constant over all the partitioning regions, leading to a trade-off or weighting parameter that is constant {\em both in space} (\ie, on the entire image domain~\cite{mumford1989optimal,ambrosio1990approximation,paragios2000geodesic,chan2001active}) {\em and in time}, \ie, during the entire course of the (typically iterative) optimization.

Neither is desirable: consider Fig.~\ref{fig:motivation}. Panels (c) and (d) show the residual and the variance, respectively, for each  region shown in (b), into which the image (a) is partitioned. Clearly, neither the residual, nor the variance (shown as a gray-level: bright is large, dark is small), are constant in space. Thus, {\em we need a spatially adapted regularization}, beyond static image features as studied in ~\cite{grady2005multilabel,joulin2012multi,lefkimmiatis2013convex,estellers2015adaptive}, or local intensity variations~\cite{dong2009multi,grasmair2009locally}. While regularization in these works is space-varying, the variation is tied to the image statistics, and therefore constant throughout the iteration. Instead, we propose a  {\em spatially-adaptive regularization scheme} that is a function of the residual, which changes during the iteration, yielding an {\em automatically annealed} schedule whereby the changes in the residual during the iterative optimization gradually guide the strength of the prior, adjusting it both in space, and in time/iteration.

To the best of our knowledge, we are the first to present an efficient scheme that uses the Huber loss for both data fidelity and regularization, in a manner that includes standard models as a special case, within a convex optimization framework. While the Huber loss~\cite{huber1964robust} has been used before for regularizations~\cite{chambolle2011first}, to the best of our knowledge we are the first to use it both in the data and regularization terms.

Furthermore, to address the phenomenon of proliferation of multiple overlapping regions that plagues most multi-label segmentation schemes, we introduce a constraint that penalizes the common area of pairwise combinations of partitions. The  constraints often used to this end are ineffective in a convex relaxation~\cite{chambolle2012convex}, which often leads to the need for user interaction~\cite{nieuwenhuis2013spatially,nieuwenhuis2014co,zemene:eccv:2016}.

To boot, we present an efficient convex optimization algorithm within in the alternating direction method of multipliers (ADMM) framework~\cite{boyd2011distributed,parikh2014proximal} with a variable splitting technique that enables us to simplify the constraint~\cite{eckstein1992douglas,wang2008new}.

\subsection{Related Work}

One of the most popular segmentation models relies on the bi-partition/piecewise-constant assumption~\cite{chan2001active}, which has been re-cast as the optimization of a convex functional based on the thresholding theorem~\cite{chan2006algorithms,bresson2007fast}. 
In a discrete graph representation, global optimization techniques are developed based on min-cut/max-flow~\cite{greig1989exact,boykov2001fast,grady2008reformulating,komodakis2007fast,komodakis2011mrf}, and there is an approach that has been applied to general multi-label problems~\cite{ishikawa2003exact,komodakis2008performance}. Continuous convex  relaxation techniques have been applied to multi-label problems in a variational framework~\cite{pock2010global,goldluecke2013tight,strekalovskiy2014convex}, where the minimization of Total Variation (TV) is performed using a primal-dual algorithm. In minimizing TV, a functional lifting technique has been applied to the multi-label problem~\cite{pock2008convex,pock2009algorithm,laude2016sublabel}.
Most convex relaxation approaches for the multi-label problems have been based on TV regularization while different data fidelity terms have been used by $L_1$ norm~\cite{unger2012joint} or $L_2$ norm~\cite{brown2010convex}. 
For the regularization term, the Huber norm has been used for TV in order to avoid undesirable staircase effects~\cite{chambolle2011first}. There have been adaptive algorithms proposed to improve the accuracy of the boundary using an edge indicator function~\cite{perona1990scale,grady2005multilabel,joulin2012multi}, generalized TV~\cite{grasmair2010anisotropic,yan2015image}, or an anisotropic structure tensor~\cite{wedel2009structure,roussos2010tensor,lefkimmiatis2013convex,estellers2015adaptive}. 
A local variation of the image intensity within a fixed size window has been also applied to consider local statistics into the regularization~\cite{dong2009multi,grasmair2009locally}, and the regularization parameter has been chosen based on the noise variance~\cite{galatsanos1992methods}.
Most adaptive regularization algorithms have considered spatial statistics that are constant during the iteration, irrespective of the residual. 

It has been known that most multi-label models suffer from inaccurate or duplicate partitioned regions when used with a large number of labels~\cite{chambolle2012convex}, which forces user interactions including bounding boxes~\cite{liu2010fast,rother2004grabcut,vicente2009joint}, contours~\cite{arbelaez2009contours,blake2004interactive}, scribbles~\cite{li2004lazy,wang2007discriminative,nieuwenhuis2013spatially,nieuwenhuis2014co,zemene:eccv:2016}, or points~\cite{bearman:eccv:2016}. 
Alternatively, side information such as depth has been applied to overcome the difficulties that stem from uncertainty in characterization of regions in the multi-label problem~\cite{shao2012interactive,diebold2015interactive,banica2015second,fu2015object,Feng_2016_CVPR}.

The words ``deep learning'' appear nowhere in this paper but this sentence: we believe there are {\em deep} connections between the dynamic data-driven regularization we have proposed and a process to design models that best exploit the informative content of the data, {\em learning} which can inform more useful models moving forward. Sec.~\ref{sec:information} is a first step in this direction.

\subsection{Contributions}

Our first contribution is a multi-label segmentation model that adapts locally, in space and time/iteration, to the (data-driven) statistics of the residual (Sec.~\ref{sec:residual-driven}). 

The second contribution is to introduce a Huber functional as a robust penalty for both the data fidelity and the regularization terms, which are turned into proximal operators of the $L_1$ norm, allowing us to conduct the optimization efficiently via Moreau-Yosida regularization (Sec.~\ref{sec:optimization}). 

Third, unlike most previous algorithms that were ineffective at dealing with a large number of labels, we introduce a constraint on mutual exclusivity of the labels, which penalizes the common area of pairwise combination of labels so that their duplicates are avoided (Sec.~\ref{sec:ConvexEnergy}).

Finally, we give an information-theoretic interpretation of our cost function, which highlights the underlying assumption and justifies the adaptive regularization scheme in a way that the classical Bayesian interpretation is not suited for the derivation of our model (Sec.~\ref{sec:information}).

We validate our model empirically in Sec.~\ref{sec:experiments}, albeit with the proviso that existing benchmarks are just one representative choice for measuring how {\em correct} a segmentation is, as hinted at in the introduction, whereas our hope is that our method will be {\em useful} in a variety of settings beyond the benchmarks themselves, for which purpose we intend to make our code available open-source upon completion of the anonymous review process.

\section{Multi-Label via Adaptive Regularization}

\subsection{General Multi-Label Segmentation} \label{sec:MultiLabel}

Let $f : \Omega \rightarrow \R$ be a real valued\footnote{Vector-valued images can also be handled, but we consider scalar for ease of exposition.} image with $\Omega \subset {\mathbb R}^2$ its spatial domain.
Segmentation aims to divide the domain $\Omega$ into a set of $n$ pairwise disjoint regions $\Omega_i$ where $\Omega = \cup_{i=1}^n \Omega_i$ and $\Omega_i \cap \Omega_j = \varnothing$ if $i \neq j$.  
The partitioning is represented by a labeling function $l : \Omega \rightarrow \Lambda$ where $\Lambda$ denotes a set of labels with $| \Lambda | = n$. The labeling function $l(x)$ assigns a label to each point $x \in \Omega$ such that $\Omega_i = \{ x \, | \, l(x) = i \}$.  
Each region $\Omega_i$ is indicated by the characteristic function $\chi_i : \Omega \rightarrow \{ 0, 1 \}$ defined by:
\begin{align} \label{eq:CharacteristicFunction}
\chi_i(x) = 
   \begin{cases}
   1 & : l(x) = i,\\
   0 & : l(x) \neq i.
   \end{cases}
\end{align}
Segmentation of the image $f(x)$ can be cast as an energy minimization problem in a variational framework, by seeking for regions $\{ \Omega_i \}$ that minimize an energy functional with respect to a set of characteristic functions $\{ \chi_i \}$:
\begin{align} \label{eq:NonConvexEnergy}
\sum_{i \in \Lambda} \left\{ \lambda \, \mathcal{D} (\chi_i) + (1 - \lambda) \, \mathcal{R} (D \chi_i) \right\}, \, \sum_{i \in \Lambda} \chi_i(x) = 1,
\end{align}
where $\mathcal{D}$ measures the data fidelity and $\mathcal{R}$ measures the regularity of $D \chi_i$, and $D$ denotes an operator for the distributional derivative.
The trade-off between the data fidelity $\mathcal{D}$ and the regularization $\mathcal{R}$ is controlled by the relative weighting parameter $\lambda \in [0, 1]$. 
A simple data fidelity can be designed to measure the homogeneity of the image intensity based on a piecewise constant model with an additional noise process: $f(x) = c_i + \eta_i(x), \, x \in \Omega_i$ where $c_i \in \R$ and $\eta_i(x)$ is assumed to follow a certain distribution independently in $i \in \Lambda$ with a specified parameter $\sigma \in \R$.
The regularization is designed to penalize the boundary length of region $\Omega_i$ that is preferred to have a smooth boundary. 
The classical form of the data fidelity and the regularization is:
\begin{align}
\hspace{-7pt}\mathcal{D} = \int_\Omega \chi_i(x) \, | f(x) - c_i |^p \ud x, \, \mathcal{R} = \int_\Omega | \nabla \chi_i(x) | \ud x \label{eq:NonConvexTerms}
\end{align}
where $p$ is given depending on the noise distribution assumption (e.g. $p$=1 for Laplacian and $p$=2 for Gaussian).
The control parameter $\lambda$ in~\eqref{eq:NonConvexEnergy} is related to the parameter $\sigma$ of the noise distribution leading to be constant for all $i$ since $\sigma$ is assumed to be the same for all $i$.
The energy formulation in~\eqref{eq:NonConvexEnergy} with the terms defined in~\eqref{eq:NonConvexTerms} is non-convex due to the integer constraint of the characteristic function $\chi_i(x) \in \{0, 1\}$, and the control parameter $\lambda$ is given to be constant in $\Omega$ for all $i$ due to the assumption that the associated parameter $\sigma$ with the noise distribution is constant.
We present a convex energy formulation with a novel constraint on the set of partitioning functions in the following section.

\subsection{Energy with Mutually Exclusive Constraint} \label{sec:ConvexEnergy}

We derive a convex representation of the energy functional in~\eqref{eq:NonConvexEnergy} following classical convex relaxation methods presented in~\cite{chambolle2011first,chambolle2012convex} as follows:
\begin{align} \label{eq:ConvexEnergy}
&\sum_{i \in \Lambda} \left\{ \int_\Omega \lambda \, \rho(u_i(x)) \ud x + \int_\Omega (1 - \lambda) \, \gamma( \nabla u_i(x)) \ud x \right\}, \nonumber\\
&\textrm{subject to } u_i(x) \in [0, 1], \, \sum_{i \in \Lambda} u_i(x) = 1, \, \forall x \in \Omega,
\end{align}
where $\rho(u_i)$ represents the data fidelity, $\gamma(\nabla u_i)$ represents the regularization, and their relative weight is determined by $\lambda$ about which will be discussed in the following section.  
The characteristic function $\chi_i$ in~\eqref{eq:NonConvexEnergy} is replaced by a continuous function $u_i \in BV(\Omega)$ of bounded variation and its integer constraint is relaxed into the convex set $u_i \in [0, 1]$. 
In the determination of the energy functional in~\eqref{eq:ConvexEnergy}, we employ a robust penalty estimator using the Huber loss function $\phi_{\eta}$ with a threshold parameter $\eta > 0$ as defined in~\cite{huber1964robust}:
\begin{align} \label{eq:HuberFunction}
\phi_{\eta}(x) =
   \begin{cases}
   \frac{1}{2 \eta} x^2 & : |x| \le \eta,\\
   |x| - \frac{\eta}{2} & : |x| > \eta.
   \end{cases}
\end{align}
We define the data fidelity $\rho(u_i)$ and the regularization $\gamma(\nabla u_i)$ using the Huber loss function as follows:
\begin{align}
\rho(u_i(x); c_i) &\coloneqq \phi_{\eta} (f(x) - c_i) \, u_i(x), \label{eq:DataFidelityHuber}\\
\gamma(\nabla u_i(x)) &\coloneqq \phi_{\mu} (\nabla u_i(x)), \label{eq:RegularizationHuber}
\end{align}
where $c_i \in \R$ is an approximate of $f$ to estimate within the region $\Omega_i$, and the threshold parameters $\eta, \mu > 0$ are related to the selection of the distribution model depending on the residual. 
The data fidelity $\rho(u_i; c_i)$ is defined by following the piecewise image constant model, however it can be generalized to $\rho(u_i; c_i) \coloneqq - \log p_i(f(x))$ where $p_i(f(x))$ is the probability that the observation $f(x)$ fits a certain model distribution $p_i$. 
The advantage of using the Huber loss in comparison to the $L_2$ norm is that geometric features such as edges are better preserved while it has continuous derivatives whereas the $L_1$ norm is not differentiable leading to staircase artifacts.
In addition, the Huber loss enables efficient convex optimization algorithm due to its equivalence to the proximal operator of $L_1$ norm, which will be discussed in Sec.~\ref{sec:optimization}. 

The  regions are desired to be pairwise disjoint, namely $\Omega_i \cap \Omega_j = \varnothing$ if $i \neq j$, however the summation constraint $\sum_{i \in \Lambda} u_i(x) = 1, \, \forall x \in \Omega$ in~\eqref{eq:ConvexEnergy} is ineffective to this end, especially  with a large number of labels~\cite{chambolle2012convex}.
Thus, we introduce a novel constraint to penalize the common area of each pair of combinations in regions $\Omega_i$ in such a way that $\sum_{i \neq j} u_i u_j$ is minimized for all $i, j \in \Lambda$. Then, we arrive at the following energy functional with the proposed mutually exclusive constraint:
\begin{align} \label{eq:ConvexEnergyAreaConstraint}
&\sum_{i \in \Lambda} \bigg\{ \int_\Omega \lambda \, \rho(u_i(x); c_i) + \tau \Big( \sum_{i \neq j} u_j(x) \Big) u_i(x) \ud x \nonumber\\
&+ \int_\Omega (1 - \lambda) \, \gamma( \nabla u_i(x)) \ud x \bigg\}, \nonumber\\
&\textrm{subject to } u_i(x) \in [0, 1], \, \sum_{i \in \Lambda} u_i(x) = 1, \, \forall x \in \Omega,
\end{align}
where $\tau > 0$ is a weighting parameter for the constraint of the mutual exclusivity in segmenting regions, and $\lambda$ determines the trade-off between the data fidelity and the regularization. 
The optimal partitioning functions $u_i$ and the approximates $c_i$ are computed by minimizing the energy functional in~\eqref{eq:ConvexEnergyAreaConstraint} in an alternating way. The desired segmentation results are obtained by the optimal set of partitioning functions $u_i$ as given by:
\begin{align} \label{eq:ComputeLabel}
l(x) = \argmax_i u_i(x), \quad i \in \Lambda, x \in \Omega.
\end{align}
In the optimization of the energy functional in~\eqref{eq:ConvexEnergyAreaConstraint}, we propose a novel regularization scheme that is adaptively applied based on the local fit of data to the model for each label as discussed in the following section.

\subsection{Residual-Driven Regularization}
\label{sec:residual-driven}

The trade-off between the data fidelity $\rho(u_i; c_i)$ and the regularization $\gamma(\nabla u_i)$ in~\eqref{eq:ConvexEnergyAreaConstraint} is determined by $\lambda$ based on the noise distribution in the image model.
We assume that the diversity parameter in the probability density function of the residual that is defined by the difference between the observed and predicted values varies in label.
We propose a novel regularization scheme based on the adaptive weighting function $\lambda_i$ depending on the data fidelity $\rho(u_i; c_i)$ as defined by:
\begin{align}
\nu_i(x) &= \exp\left( - \frac{\rho(u_i(x); c_i)}{\beta} \right), \label{eq:exp}\\
\lambda_i(x) &= \arg\min_{\lambda} \frac{1}{2} \| \nu_i(x) - \lambda \|_2^2 + \alpha \, \| \lambda \|_1, \label{eq:lasso}
\end{align}
where $\beta > 0$ is a constant parameter that is related to the variation of the residual, and $\alpha > 0$ is a constant parameter for the sparsity regularization.
The relative weight $\lambda_i(x)$ between the data fidelity and the regularization is adaptively applied for each label $i \in \Lambda$ and space $x \in \Omega$ depending on $\nu_i(x)$ determining the local fit of data to the model.
The adaptive regularity scheme based on the weighting function $\lambda_i(x)$ is designed so that regularization is stronger when the residual is large, equivalently $\nu_i(x)$ is small, and weaker when the residual is small, equivalently $\nu_i(x)$ is large, during the energy optimization process.
We impose sparsity on the exponential measure of the negative residual $\nu_i(x)$ to obtain the weighting function $\lambda_i(x)$ as a solution of the Lasso problem~\cite{tibshirani1996regression} defined in~\eqref{eq:lasso}.
Such a solution with the identity operator as a predictor matrix can be efficiently obtained by the soft shrinkage operator $\mathcal{T}(\nu | \alpha)$~\cite{boyd2004convex}:
\begin{align}
\mathcal{T} (\nu \, | \, \alpha) &= 
\begin{cases}
\nu - \alpha & : \nu > \alpha\\
0 & : \| \nu \|_1 \le \alpha\\
\nu + \alpha & : \nu < -\alpha\\
\end{cases}
\label{eq:shrink}
\end{align}
leading to the solution $\lambda_i(x) = \mathcal{T}(\nu_i(x) | \alpha)$.
The Lagrange multiplier $\alpha > 0$ in the Lasso problem in~\eqref{eq:lasso} where $0 < \nu_i \le 1$ restricts the range of $\lambda_i$ to be $[0, 1-\alpha]$ so that the regularization is imposed everywhere, which leads to well-posedness even if $\rho(u_i; c_i) = 0$.
The constant $\alpha$ is related to the overall regularity on the entire domain.
The final energy functional for our multi-label segmentation problem with the adaptive regularity scheme reads:
\begin{align} \label{eq:ConvexEnergyAdaptive}
&\sum_{i \in \Lambda} \bigg\{ \int_\Omega \lambda_i(x) \, \rho(u_i(x); c_i) + \tau \Big( \sum_{i \neq j} u_j(x) \Big) u_i(x) \ud x \nonumber\\
&+ \int_\Omega (1 - \lambda_i(x)) \, \gamma( \nabla u_i(x)) \ud x \bigg\}, \nonumber\\
&\textrm{subject to } u_i(x) \in [0, 1], \, \sum_{i \in \Lambda} u_i(x) = 1, \, \forall x \in \Omega,
\end{align}
where $\lambda_i$ is obtained as the solution of~\eqref{eq:lasso}.
The optimization algorithm to minimize the energy functional in~\eqref{eq:ConvexEnergyAdaptive} is presented in Sec.~\ref{sec:optimization} and the supplementary material.


\subsection{Information-Theoretic Interpretation}
\label{sec:information}

The energy functional for our basic model incorporating the adaptive regularization, simplified after removing auxiliary variables used in the optimization and the additional constraint on the mutual exclusivity of regions, consists of a point-wise sum, which could be interpreted probabilistically by assuming that the image $f$ is a sample from an IID random field. Under that interpretation, we have that $\rho(u_i(x)) = - \log p(u(x) | f(x), c_i)$,  and $\gamma(\nabla u_i(x)) = - \log p(u(x) | c_i)$. For simplicity, we indicate those as $-\log p(u | f)$ and $-\log p(u)$ respectively, and even further $p \doteq p(u|f)$ and $q \doteq p(u)$. Then  $\lambda_i(x) \propto p$, and the overall cost function \eqref{eq:ConvexEnergy} to be {\em minimized}, $\int \lambda \rho + (1 - \lambda) \gamma dx$, can be written as
\begin{equation}
\mathcal{E}^{\rm min}_{p,q} = - {\mathbb E}\left(p \, \log \frac{p}{q}\right) + {\mathbb E}(\log q),
\end{equation}
where the expectation $\mathbb{E}$ is the sum with respect to the values taken by $u$ and $f$ on $x$. The first term is the Kullback-Leibler divergence between the chosen prior and the data-dependent posterior. The second term is constant once the prior is chosen. Therefore, the model chosen for the posterior is the one that {\em maximizes} the divergence between prior and posterior,
where the influence of the prior $q$ wanes as the solution becomes an increasingly better fit of the data, without the need for manual tuning of the annealing, and without the undesirable bias of the prior on the final solution. The model chosen is therefore the one that, for a fixed prior, selects the posterior to be as divergent as possible, so as to make the data as informative as possible (in the sense of uncertainty reduction).

Compare this with the usual Bayesian interpretation of variational segmentation, whereby the function to be maximized is 
\begin{equation}
\mathcal{F}^{\rm min}_{p,q} = \int \log p + \beta \, \log q \ud x,
\end{equation}
for some fixed $\beta$ and a prior $q$ whose influence does not change with the data. If we wanted it to change, we would have to introduce an annealing parameter $\lambda$, so 
\begin{equation}
\mathcal{F}^{\rm max}_{p,q} = p^\lambda \, q^{(1-\lambda)},
\end{equation}
with no guidance from Bayesian theory on how to choose it or schedule it. Clearly choosing $\lambda = p$ yields a form that is not easily interpreted within Bayesian inference. Thus the information-theoretic setting provides us with guidance on how to choose the parameter $\lambda$, whereby the cost function is given as:
\begin{equation}
\mathcal{E}^{\rm max}_{p,q} = {\mathbb{KL}}(p \, || \, q),
\end{equation}
where $\mathbb{KL}$ denotes the Kullback-Leibler divergence. We obtain $\lambda = p$ and consequently an automatic, data-driven annealing schedule.

\section{Energy Optimization} \label{sec:optimization}

In this section, we present an efficient convex optimization algorithm in the framework of alternating direction method of multipliers (ADMM)~\cite{boyd2011distributed,parikh2014proximal}. The detailed derivations of our optimization algorithm are further provided in the supplementary material.

The energy~\eqref{eq:ConvexEnergyAdaptive}, that is convex in $u_i$ with fixed $c_i$, is minimized with respect to the partitioning functions $u_i$ and the intensity estimates $c_i$ in an alternating way. 
We modify the energy functional in~\eqref{eq:ConvexEnergyAdaptive} using variable splitting~\cite{courant1943variational,eckstein1992douglas,wang2008new} introducing a new variable $v_i$ with the constraint $u_i = v_i$ as follows:
\begin{align} \label{eq:ConvexEnergySplit}
&\sum_{i \in \Lambda} \bigg\{ \int_\Omega \lambda_i \, \rho(u_i; c_i) + \tau \Big( \sum_{i \neq j} u_j \Big) u_i \ud x \nonumber\\
&+ \int_\Omega (1 - \lambda_i) \, \gamma( \nabla v_i ) \ud x + \frac{\theta}{2} \| u_i - v_i + y_i \|_2^2 \bigg\}, \nonumber\\
&\textrm{subject to } u_i(x) \ge 0, \, \sum_{i \in \Lambda} v_i(x) = 1, \, \forall x \in \Omega,
\end{align}
where $y_i$ is a dual variable for each equality constraint $u_i = v_i$, and $\theta > 0$ is a scalar augmentation parameter.
The original constraints $u_i \in [0, 1]$ and $\sum_i u_i = 1$ in~\eqref{eq:ConvexEnergyAdaptive} are decomposed into the simpler constraints $u_i \ge 0$ and $\sum_i v_i = 1$ in~\eqref{eq:ConvexEnergySplit} by variable splitting $u_i = v_i$.
In the computation of the data fidelity and the regularization, we employ a robust estimator using the Huber loss function defined in~\eqref{eq:HuberFunction}. 
An efficient procedure can be performed to minimize the Huber loss function $\phi_{\eta}$ following the equivalence property of Moreau-Yosida regularization of a non-smooth function $| \cdot |$ as defined by~\cite{moreau1965proximite,opac-b1133911}:
\begin{align} \label{eq:MoreauYosida}
\phi_{\eta}(x) = \inf_r \left\{ | r | + \frac{1}{2 \eta} (x - r)^2 \right\},
\end{align}
which replaces the data fidelity $\rho(u_i; c_i)$ and the regularization $\gamma(\nabla v_i)$ in~\eqref{eq:ConvexEnergySplit} with the regularized forms $\rho(u_i; c_i, r_i)$ and $\gamma(\nabla v_i; z_i)$, respectively as follows:
\begin{align} 
\rho(u_i; c_i, r_i) &= \left( | r_i | + \frac{1}{2 \eta} (f - c_i - r_i)^2 \right) u_i, \label{eq:DataFidelityMoreauYosida}\\
\gamma(\nabla v_i; z_i) &=  \| z_i \|_1 + \frac{1}{2 \mu} \| \nabla v_i - z_i \|^2_2, \label{eq:RegularizationMoreauYosida}
\end{align}
where $r_i$ and $z_i$ are the auxiliary variables to be minimized in alternation.
The constraints on $u_i$ and $v_i$ in~\eqref{eq:ConvexEnergySplit} can be represented by the indicator function $\delta_A(x)$ of a set $A$ defined by:
\begin{align} \label{eq:DeltaFunction}
\delta_A(x) &= 
   \begin{cases}
   0 & : x \in A\\
   \infty & : x \notin A\\
   \end{cases}
\end{align}
The constraint $u_i \ge 0$ is given by $\delta_A(u_i)$ where $A = \{ x | x \ge 0 \}$, and the constraint $\sum_i v_i = 1$ is given by $\delta_B( \{ v_i \} )$ where $B = \left\{ \{x_i\} | \sum_i x_i = 1 \right\}$.
The Moreau-Yosida regularization for the Huber loss function and the constraints represented by the indicator functions lead to the following unconstrained energy functional $\mathcal{L}_i$ for label $i$:
\begin{align} \label{eq:EnergyUnconstrainedLabel}
&\mathcal{L}_i = \int_\Omega \lambda_i \, \rho(u_i; c_i, r_i) + \tau \Big( \sum_{i \neq j} u_j \Big) u_i \ud x + \delta_A(u_i) \nonumber\\
&+ \int_\Omega (1 - \lambda_i) \, \gamma( \nabla v_i; z_i ) \ud x + \frac{\theta}{2} \| u_i - v_i + y_i \|_2^2,
\end{align}
and the final unconstrained energy functional $\mathcal{E}$ reads:
\begin{align} \label{eq:EnergyUnconstrained}
\mathcal{E}(\{u_i, v_i, y_i, c_i, r_i, z_i\}) = \sum_{i \in \Lambda} \mathcal{L}_i + \delta_B(\{v_i\}).
\end{align}
The optimal set of partitioning functions $\{ u_i \}$ is obtained by minimizing the energy functional $\mathcal{E}$ using ADMM, minimizing the augmented Lagrangian $\mathcal{L}_i$ in~\eqref{eq:EnergyUnconstrainedLabel} with respect to the variables $u_i, v_i, c_i, r_i, z_i$, and applying a gradient ascent scheme with respect to the dual variable $y_i$ followed by the update of the weighting function $\lambda_i$ and the projection of $\{ v_i \}$ onto the set $B$ in~\eqref{eq:EnergyUnconstrained}. 
The alternating optimization steps using ADMM are presented in Algorithm~\ref{alg:admm}, where $k$ is the iteration counter. 
%
%
\begin{algorithm}[tb]
\caption{The ADMM updates for minimizing~\eqref{eq:EnergyUnconstrained}}
\label{alg:admm}
\begin{algorithmic}
\State 
\textbf{for} each label $i \in \Lambda$ \textbf{ do}
\begin{flalign}
\;\; c_i^{k+1} & \coloneqq \argmin_{c} \rho( u_i^k; c, r_i^k ) & \label{step:c}\\
\;\; r_i^{k+1} & \coloneqq \argmin_{r} \rho( u_i^k; c_i^{k+1}, r ) & \label{step:r}\\
\;\; \nu_i^{k+1} & \coloneqq \exp\left( - \frac{\rho(u_i^k; c_i^{k+1}, r_i^{k+1})}{\beta} \right) & \label{step:nu}\\
\;\; \lambda_i^{k+1} & \coloneqq \argmin_{\lambda} \frac{1}{2} \| \nu_i^{k+1} - \lambda \|_2^2 + \alpha \| \lambda \|_1 & \label{step:l}\\
\;\; z_i^{k+1} & \coloneqq \argmin_z \gamma(\nabla v_i^k; z) & \label{step:z}\\
\;\; u_i^{k+1} & \coloneqq \argmin_{u} \int_\Omega \lambda_i^{k+1} \rho(u; c_i^{k+1}, r_i^{k+1}) \ud x + \delta_A(u) & \nonumber\\
& \hspace{-6pt} + \int_\Omega \tau \Big( \sum_{i \neq j} u_j \Big) u \ud x + \frac{\theta}{2} \| u - v_i^k + y_i^k \|_2^2  & \label{step:u}\\
\;\; \tilde{v}_i^{k+1} & \coloneqq \argmin_{v} \int_\Omega \left( 1-\lambda_i^{k+1} \right) \gamma(\nabla v; z_i^{k+1}) \ud x & \nonumber\\
& \hspace{-6pt} + \frac{\theta}{2} \| u_i^{k+1} - v + y_i^k \|_2^2 & \label{step:v}\\
y_i^{k+1} & \coloneqq y_i^k + u_i^{k+1} - v_i^{k+1} & \label{step:y}
\end{flalign}
\textbf{end for}
\begin{flalign}
\{ v_i^{k+1} \} & \coloneqq \Pi_B \left( \{ \tilde{v}_i^{k+1} \} \right) & \label{step:v:projection}
\end{flalign}
\end{algorithmic}
\end{algorithm}
%
%
The technical details regarding the optimality conditions and the optimal solutions for the update of variable at each step in Algorithm~\ref{alg:admm} are provided in the supplementary material.
\section{Experimental Results} \label{sec:experiments}

We demonstrate the effectiveness and robustness of our proposed algorithm for the multi-label segmentation problem using the images in the Berkeley segmentation dataset~\cite{arbelaez2011contour} and synthetic images.
The numerical experiments are designed to demonstrate the robustness of the energy functional that uses a Huber function as a penalty estimator, the effectiveness of the mutually exclusive constraint on the partitioning functions, and the effectiveness of the adaptive regularity scheme based on the local fit of data to the model. Note that we use a random initialization for the initial labeling function for all the algorithms throughout the experiments.

{\bf Robustness of the Huber-Huber (${\bf H}^2$) Model:}\\
%
%
\def\fh{87pt}
\begin{figure}
\centering
\begin{tabular}{c@{}c}
\includegraphics[totalheight=\fh]{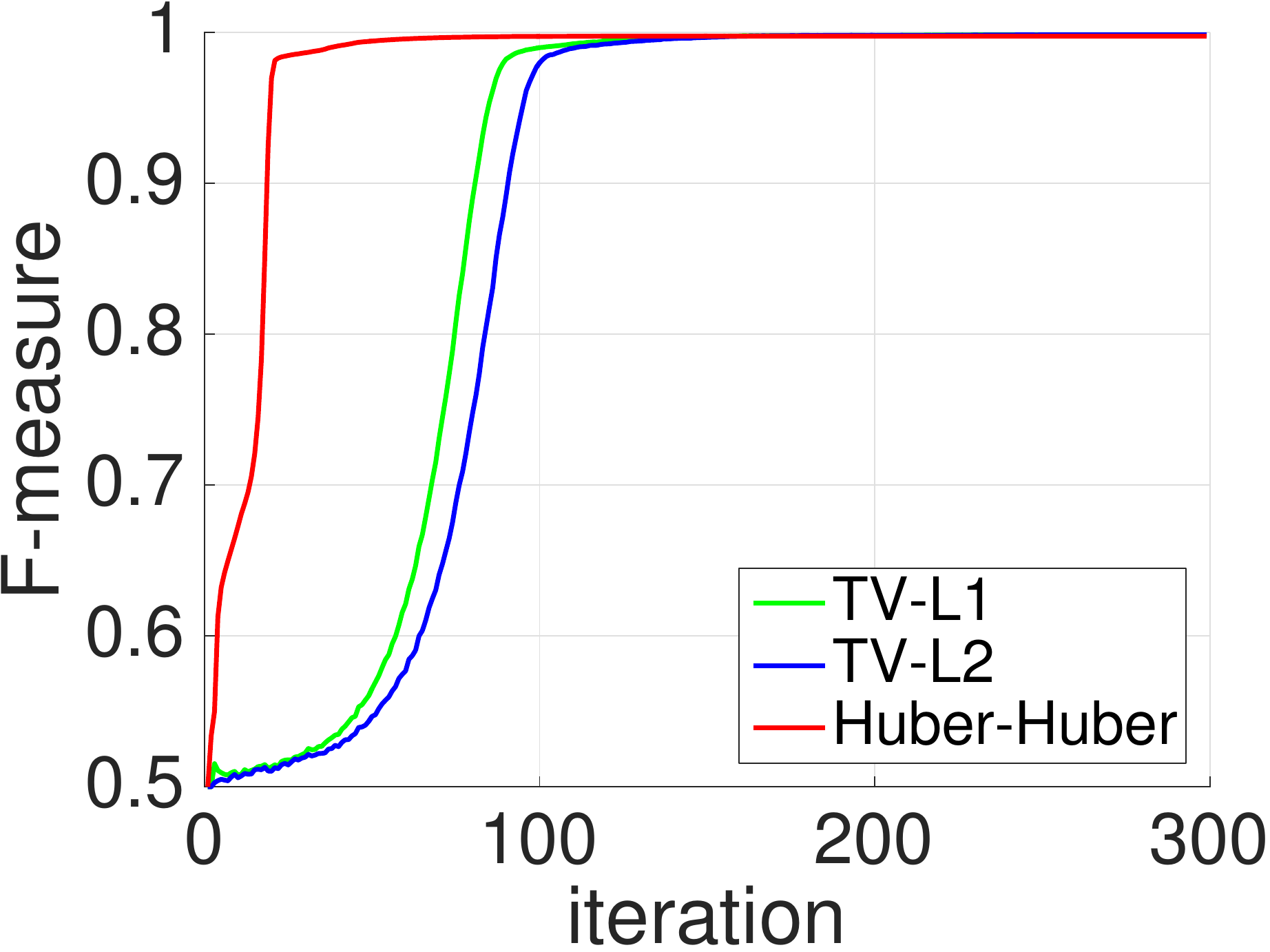} &
\includegraphics[totalheight=\fh]{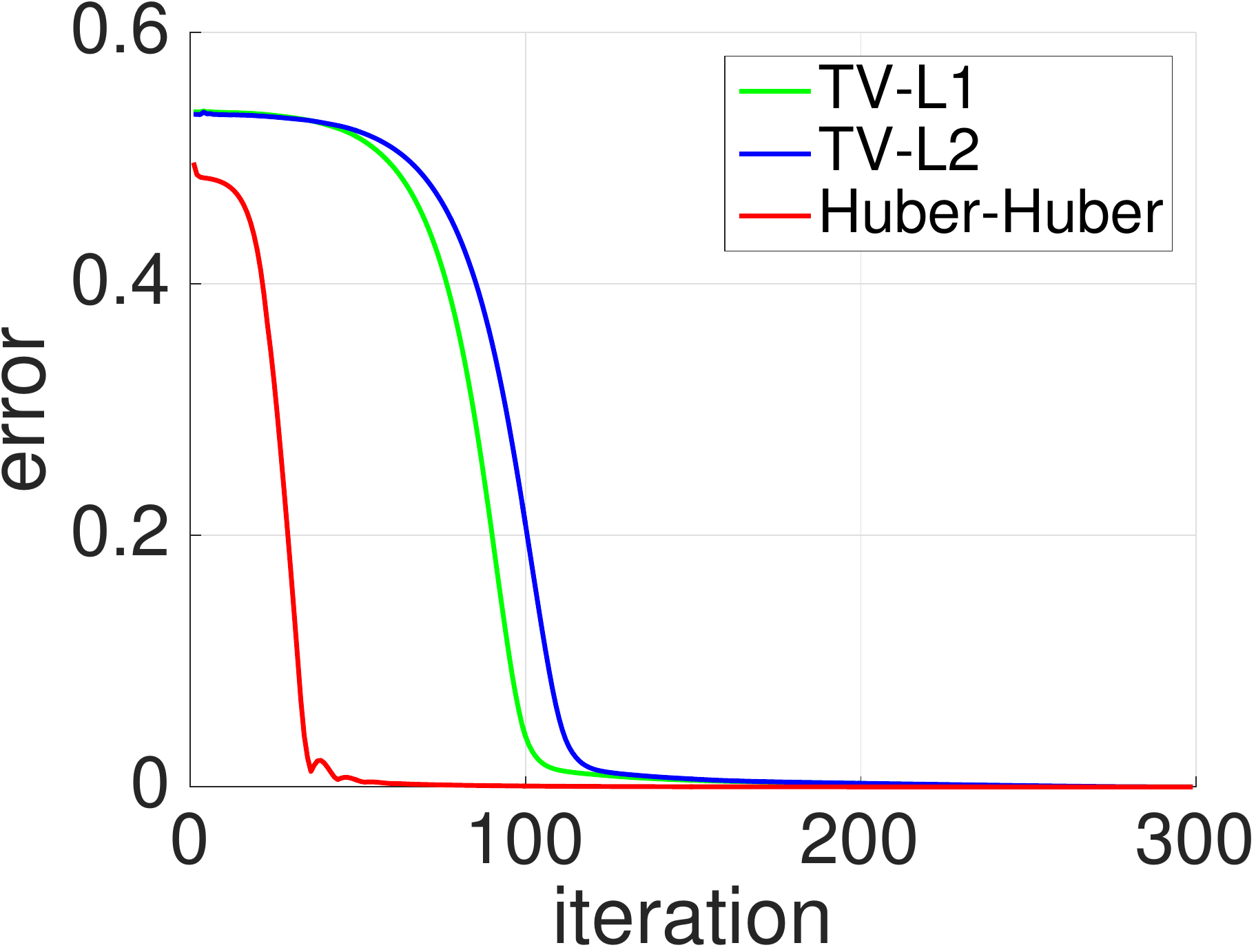}\\
(a) Accuracy & (b) Error
\end{tabular}
\caption{Quantitative Comparison of the different energy models using the F-measure (left) and residual (right) as a function of the number of iterations. We compare the popular TV-$L_1$ and TV-$L_2$ approaches to our ${\rm H}^2$ model, which is more accurate (left) and also converges faster (right) for a bi-partitioning problem on the images suited for the bi-partitioning image model.}
\label{fig:compare:model:accuracy}
\end{figure}
%
%
We empirically demonstrate the robustness of the proposed energy functional that uses a robust Huber loss function for both the data fidelity and the regularization.
We consider a bi-partitioning segmentation on the images that are suited for a bi-partitioning image model in the Berkeley dataset in order to demonstrate the robustness of the proposed ${\rm H}^2$ model in comparison to TV-$L_1$ and TV-$L_2$ ignoring the effect of the constraint on the common area of the pairwise region combinations. 
We apply the ADMM optimization algorithm for our model with a variable splitting technique and the primal-dual algorithm is applied for TV-$L_1$ and TV-$L_2$ models.
It is shown that our model yields better and faster results as shown in Fig.~\ref{fig:compare:model:accuracy} where (a) F-measure and (b) error are presented for each iteration. The parameters for each algorithm are chosen fairly so accuracy and the convergence rate are optimized. The experimental results indicate that the proposed ${\rm H}^2$ model with the presented optimization algorithm has a potential to be applied for a variety of imaging tasks.   

{\bf Effectiveness of Mutually Exclusive Constraint:}\\
%
%
\setlength{\fboxsep}{0pt}
\setlength{\fboxrule}{0pt}
\def\fh{52pt}
\def\fw{70pt}
\def\sp{5pt}
\begin{figure*}
\centering
\begin{tabular}{c@{}c@{}c@{}c@{}c@{}c@{}c@{}c}
\fbox{\parbox[b][\fh][c]{\fw}{\# of regions = 5\\\# of labels $\;$ = 4}} &
\includegraphics[trim={20pt 20pt 20pt 20pt},clip,totalheight=\fh]{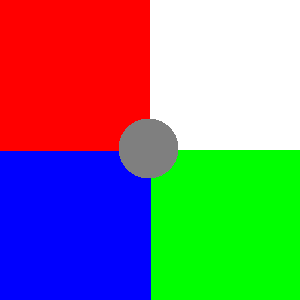} \hspace{\sp} & 
\includegraphics[trim={20pt 20pt 20pt 20pt},clip,totalheight=\fh]{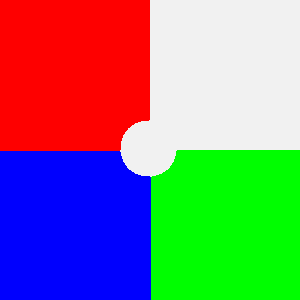} \hspace{\sp} &  
\includegraphics[trim={20pt 20pt 20pt 20pt},clip,totalheight=\fh]{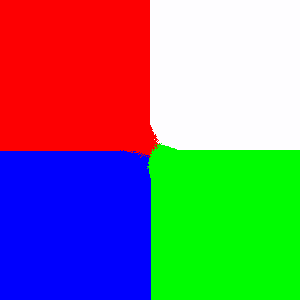} \hspace{\sp} &  
\includegraphics[trim={20pt 20pt 20pt 20pt},clip,totalheight=\fh]{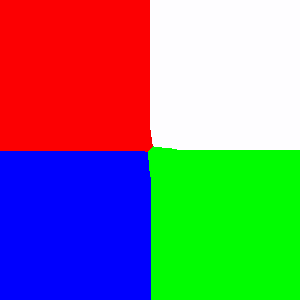} \hspace{\sp} &  
\includegraphics[trim={20pt 20pt 20pt 20pt},clip,totalheight=\fh]{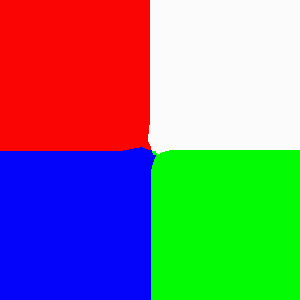} \hspace{\sp} &  
\includegraphics[trim={20pt 20pt 20pt 20pt},clip,totalheight=\fh]{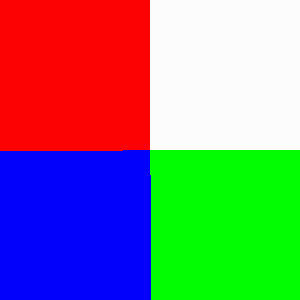} \hspace{\sp} & 
\includegraphics[trim={20pt 20pt 20pt 20pt},clip,totalheight=\fh]{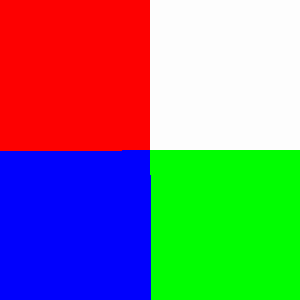} \\
\fbox{\parbox[b][\fh][c]{\fw}{\# of regions = 7\\\# of labels $\;$ = 6}} &
\includegraphics[trim={20pt 20pt 20pt 20pt},clip,totalheight=\fh]{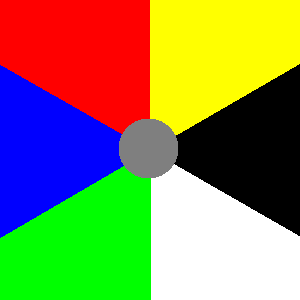} \hspace{\sp} &
\includegraphics[trim={20pt 20pt 20pt 20pt},clip,totalheight=\fh]{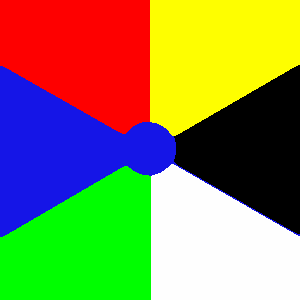} \hspace{\sp} &
\includegraphics[trim={20pt 20pt 20pt 20pt},clip,totalheight=\fh]{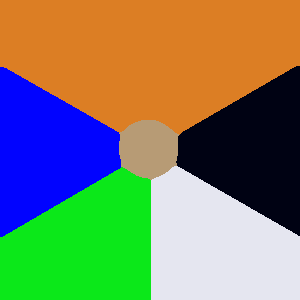} \hspace{\sp} &
\includegraphics[trim={20pt 20pt 20pt 20pt},clip,totalheight=\fh]{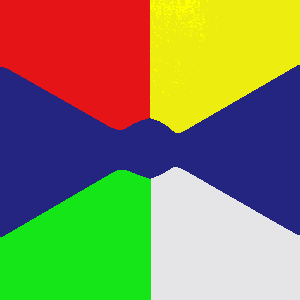} \hspace{\sp} &
\includegraphics[trim={20pt 20pt 20pt 20pt},clip,totalheight=\fh]{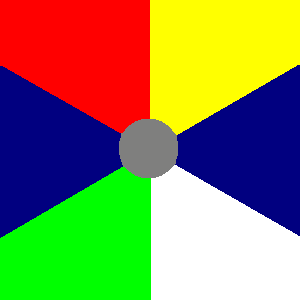} \hspace{\sp} &
\includegraphics[trim={20pt 20pt 20pt 20pt},clip,totalheight=\fh]{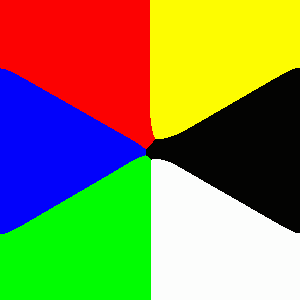} \hspace{\sp} &
\includegraphics[trim={20pt 20pt 20pt 20pt},clip,totalheight=\fh]{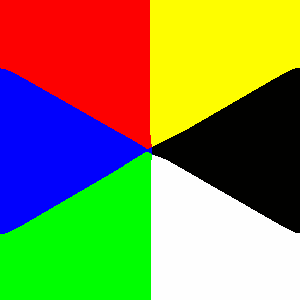} \\
\fbox{\parbox[b][\fh][c]{\fw}{\# of regions = 9\\\# of labels $\;$ = 8}} &
\includegraphics[trim={20pt 20pt 20pt 20pt},clip,totalheight=\fh]{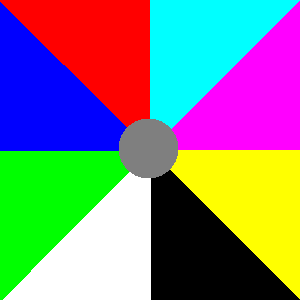} \hspace{\sp} &
\includegraphics[trim={20pt 20pt 20pt 20pt},clip,totalheight=\fh]{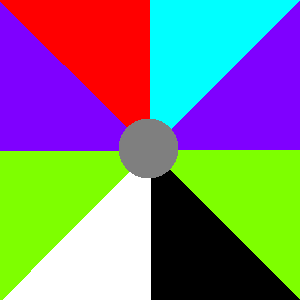} \hspace{\sp} &
\includegraphics[trim={20pt 20pt 20pt 20pt},clip,totalheight=\fh]{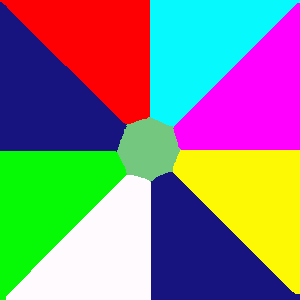} \hspace{\sp} &
\includegraphics[trim={20pt 20pt 20pt 20pt},clip,totalheight=\fh]{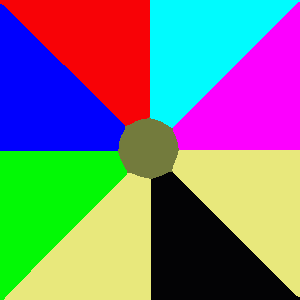} \hspace{\sp} &
\includegraphics[trim={20pt 20pt 20pt 20pt},clip,totalheight=\fh]{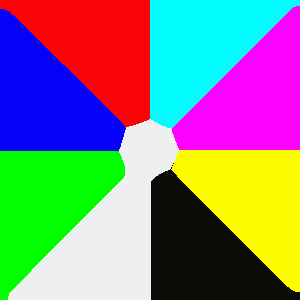} \hspace{\sp} &
\includegraphics[trim={20pt 20pt 20pt 20pt},clip,totalheight=\fh]{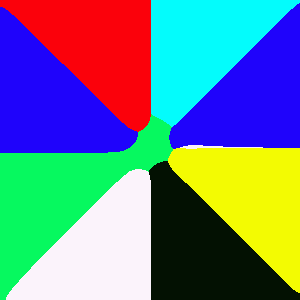} \hspace{\sp} &
\includegraphics[trim={20pt 20pt 20pt 20pt},clip,totalheight=\fh]{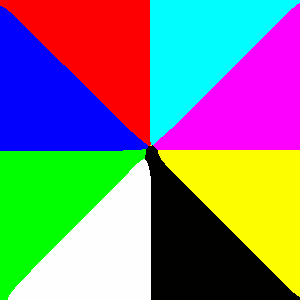} \\
& (a) Input \hspace{\sp} & (b) FL~\cite{sundaramoorthi2014fast} \hspace{\sp} & (c) TV~\cite{zach2008fast} \hspace{\sp} & (d) VTV~\cite{lellmann:continuous:siam:2011} \hspace{\sp} & (e) PC~\cite{chambolle2012convex} \hspace{\sp} & (f) our $\rm{H}^2$ \hspace{\sp} & (g) our full $\rm{H}^2$
\end{tabular}
\caption{Qualitative comparison for the {\em junction test} with increasing number of regions (5, 7, 9 from top to bottom). The number of labels is fixed at one-minus the true one (4, 6, 8 respectively), forcing the algorithm to fill in one of the regions. This test is reflective of the ability of the prior to capture the structure of the image in the presence of missing data. We compare multiple models, as indicated in the legend, all of which degrade with the number of regions; ours shows consistently better performance, as indicated by more regular in-painting (g).}
\label{fig:compare:model:constraint}
\end{figure*}
%
%
We demonstrate the effectiveness of our proposed model with a constraint of the mutual exclusivity in comparison with other multi-label segmentation algorithms.
We qualitatively compare the segmentation results with different number of labels on the classical {\em junction test} (cf. Fig.~12 of~\cite{chambolle2011first} or Fig.~5-8 of~\cite{chambolle2012convex}), whereby the number of labels is fixed to one-less than the number of regions in the input image. The algorithm is then forced to {\em ``inpaint''} the central disc with labels of surrounding regions. 
The segmentation results on the junction prototype images with different number of regions are shown in Fig.~\ref{fig:compare:model:constraint} where the input junction images have 5 (top), 7 (middle), 9 (bottom) regions as shown in (a). 
We compare (f) our Huber-Huber ($\rm{H}^2$) model without the mutual exclusivity constraint in~\eqref{eq:ConvexEnergy} and (g) our full $\rm{H}^2$ model with the constraint in~\eqref{eq:ConvexEnergyAreaConstraint} to the algorithms including: (b) fast-label (FL)~\cite{sundaramoorthi2014fast}, (c) convex relaxation based on Total Variation using the primal-dual (TV)~\cite{zach2008fast}, (d) vectorial Total Variation using the Dogulas-Rachford (VTV)~\cite{lellmann:continuous:siam:2011}, (e) paired calibration (PC)~\cite{chambolle2012convex}.
This experiment is particularly designed to demonstrate a need for the constraint of the mutual exclusivity, thus the input images are made to be suited for a precise piecewise constant model so that the underlying image model of the algorithm under comparison is relevant. The illustrative results are presented in Fig.~\ref{fig:compare:model:constraint} where the performance of most algorithms degrades as the number of regions increases (top-to-bottom), while our algorithm yields consistently better results.

{\bf Effectiveness of Adaptive Regularity:}\\
%
%
We empirically demonstrate the effectiveness of our proposed adaptive regularization using a representative synthetic image with four regions, each exhibiting spatial statistics of different dispersion, in Fig.~\ref{fig:compare:regularity} (a). The artificial noises are added to the white background, the red rectangle on the left, the green rectangle on the middle, and the blue rectangle on the right with increasing degree of noises in order.

To preserve sharp boundaries, one has to manually choose a large-enough $\lambda$ so that regularization is small; however, large intensity variance in some of the data yields undesirably irregular boundaries between regions, with red and blue scattered throughout the middle and right rectangles (d), all of which however have sharp corners. On the other hand, to ensure homogeneity of the regions, one has to crank up regularization (small $\lambda$), resulting in a biased final solution where corners are rounded (e), even for regions that would allow fine-scale boundary determination (red). Our approach with the adaptive regularization (b), however, naturally finds a solution with a sharp boundary where the data term supports it (red), and let the regularizer weigh-in on the solution when the data is more uncertain (blue). The zoom in images for the marked regions in (b) and (e) are shown in (c) and (f), respectively in order to highlight the geometric property of the solution around the corners.
%
%
\def\fh{75pt}
\def\sp{10pt}
\begin{figure}
\centering
\begin{tabular}{c@{}c@{}c}
\includegraphics[totalheight=\fh]{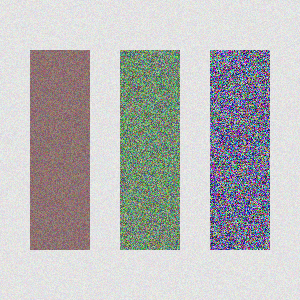} & 
\includegraphics[totalheight=\fh]{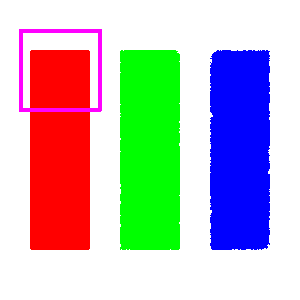} &
\includegraphics[totalheight=\fh]{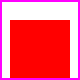} \\
(a) Input & (b) Ours (adaptive) & (c) Zoom in of (b)\\
\includegraphics[totalheight=\fh]{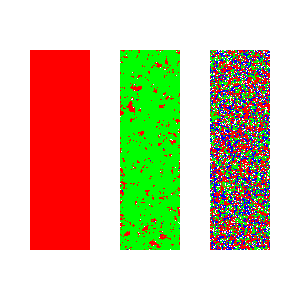} &  
\includegraphics[totalheight=\fh]{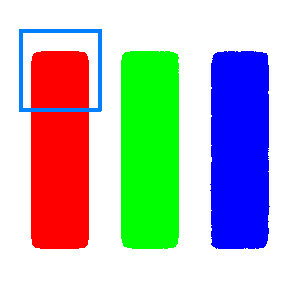} &
\includegraphics[totalheight=\fh]{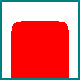} \\
(d) Small (global) & (e) Large (global) & (f) Zoom in of (e)
\end{tabular}
\caption{Qualitative comparison illustrating the pitfalls of a constant, non-adaptive, regularizer and its bias on the final solution. The image (a) has three regions with sharp boundaries/corners, and varying amount of spatial variability. Using a small amount of regularization (d) (large $\lambda$) yields sharp boundaries, but irregular partition into three regions (red, green and blue), with different labels dispersed throughout the center and right rectangles. Using a large regularizer weight (e) (small $\lambda$) yields homogeneous regions, but the boundaries are blurred out and the corners rounded, even in the left region (red). Our method (b) is driven by the data where possible (left region, sharp boundaries) and let the regularizer weigh-in where the data is more uncertain (right region, rounded boundaries).}
\label{fig:compare:regularity}
\end{figure}
%

{\bf Multi-Label Segmentation on Real Images:}\\
%
%
\def\fw{68pt}
\def\sp{1pt}
\def\dexp{exp5}
\def\sa{62096}
\def\sal{4}
\def\sb{124084}
\def\sbl{3}
\def\sc{15062}
\def\scl{4}
\def\sd{238011}
\def\sdl{3}
\def\se{12003}
\def\sel{3}
\def\sf{29030}
\def\sfl{5}
\begin{figure*}
\centering
\begin{tabular}{c@{}c@{}c@{}c@{}c@{}c}
\includegraphics[width=\fw]{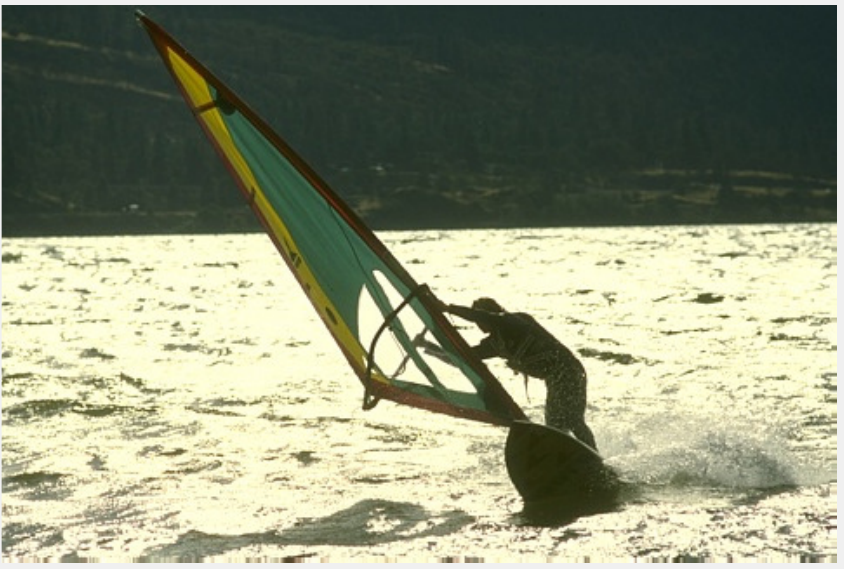} \hspace{\sp} & 
\includegraphics[width=\fw]{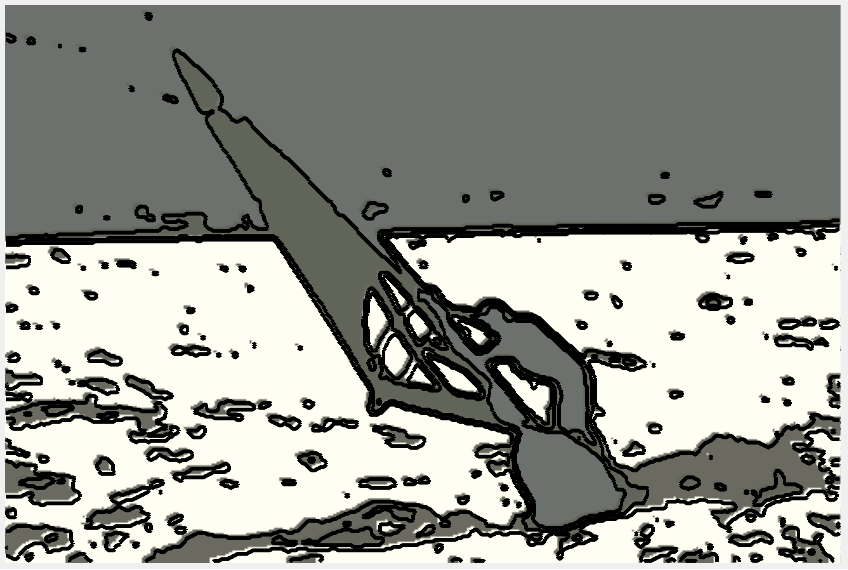} \hspace{\sp} &  
\includegraphics[width=\fw]{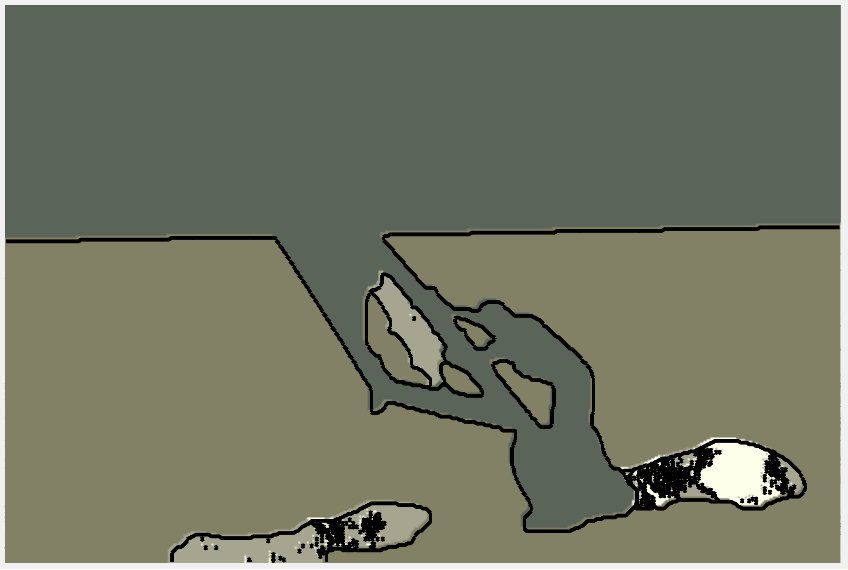} \hspace{\sp} &  
\includegraphics[width=\fw]{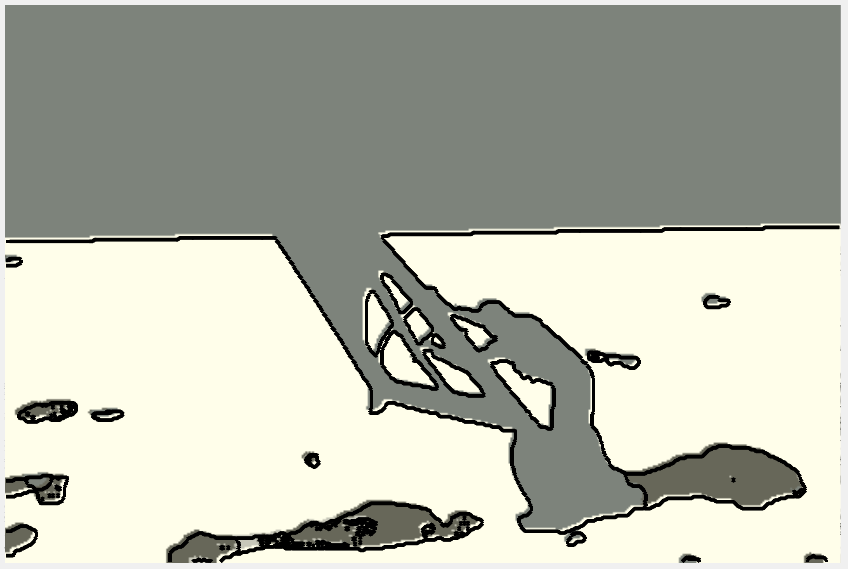} \hspace{\sp} &  
\includegraphics[width=\fw]{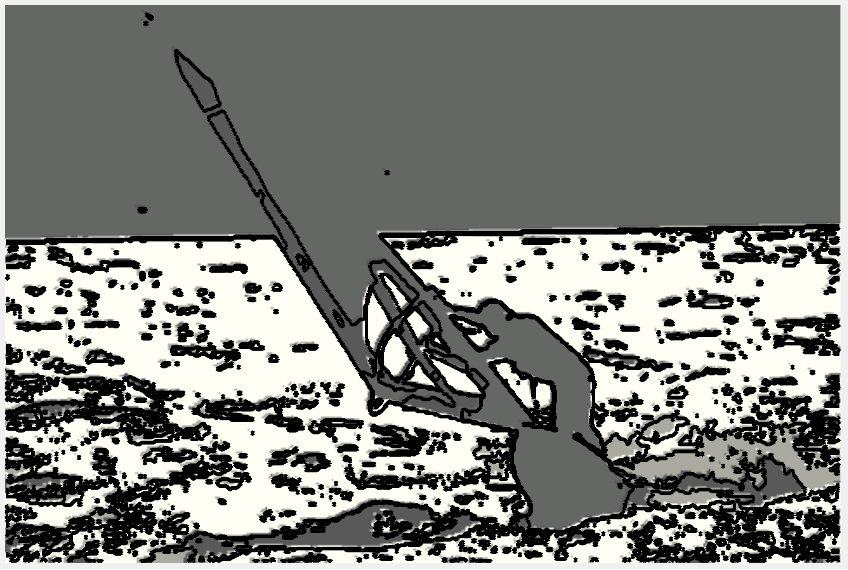} \hspace{\sp} &  
\includegraphics[width=\fw]{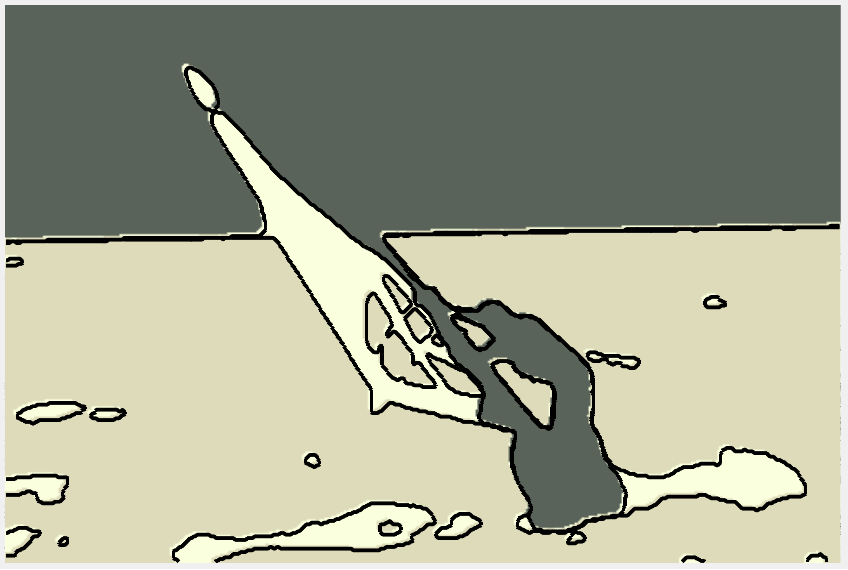} \\
\includegraphics[width=\fw]{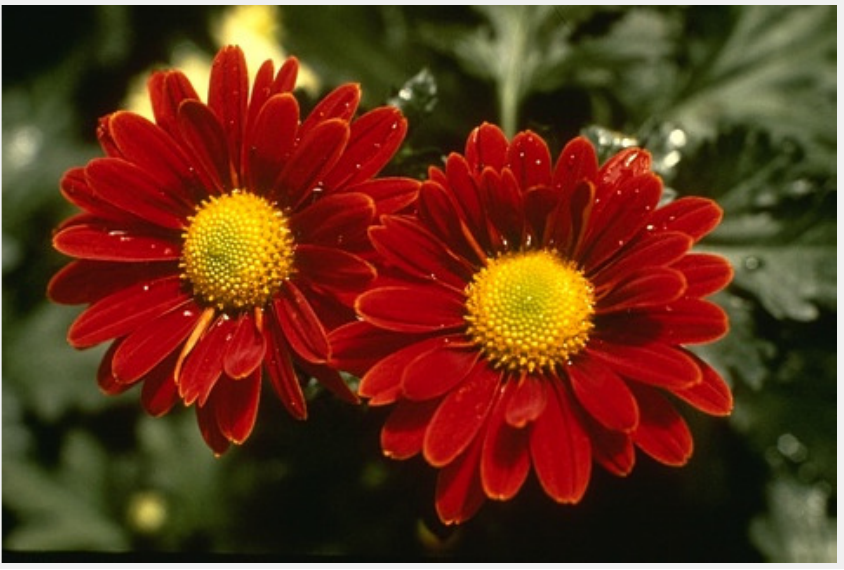} \hspace{\sp} & 
\includegraphics[width=\fw]{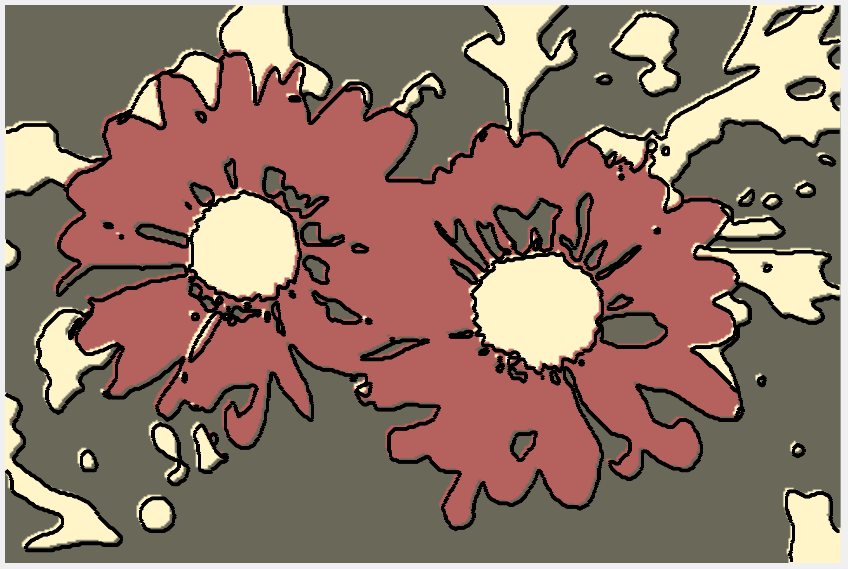} \hspace{\sp} &  
\includegraphics[width=\fw]{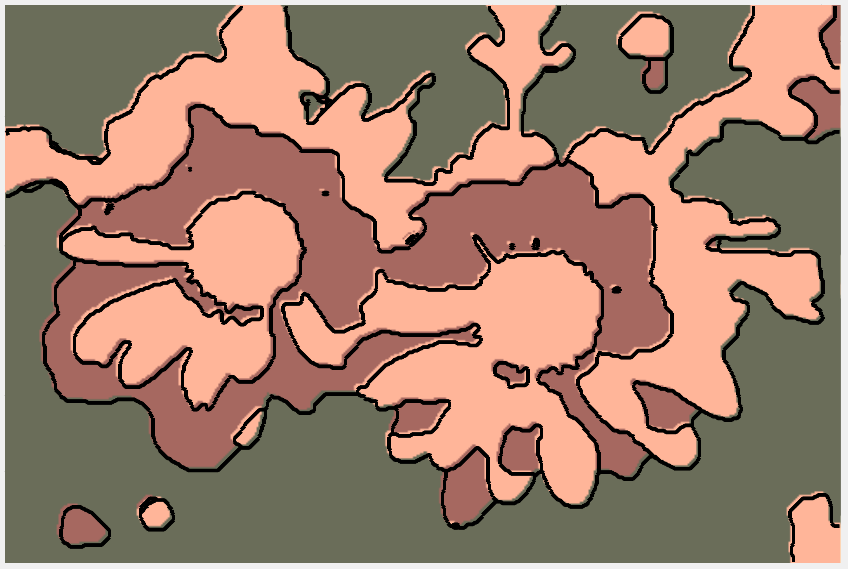} \hspace{\sp} &  
\includegraphics[width=\fw]{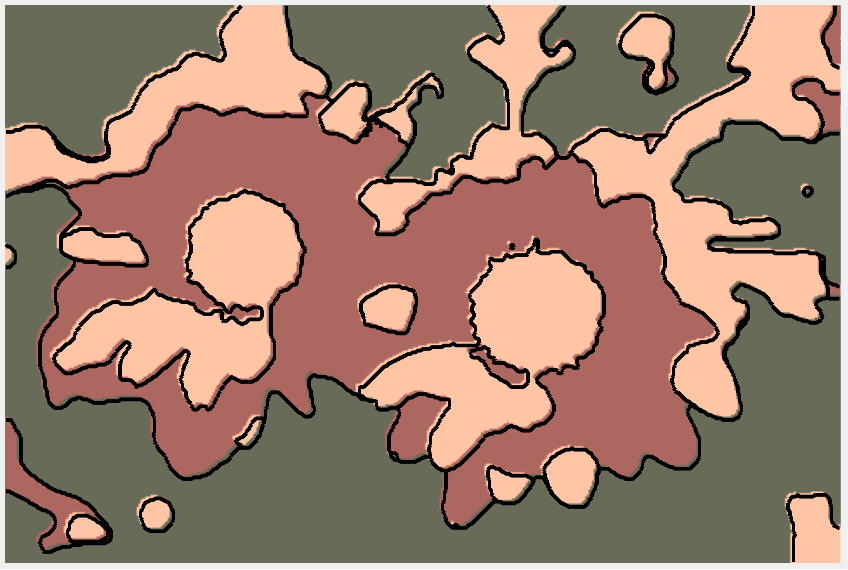} \hspace{\sp} &  
\includegraphics[width=\fw]{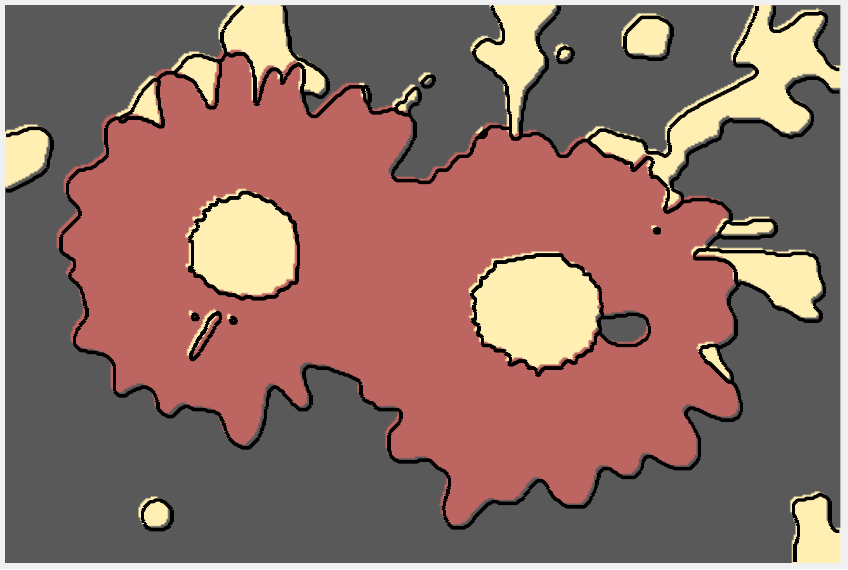} \hspace{\sp} &  
\includegraphics[width=\fw]{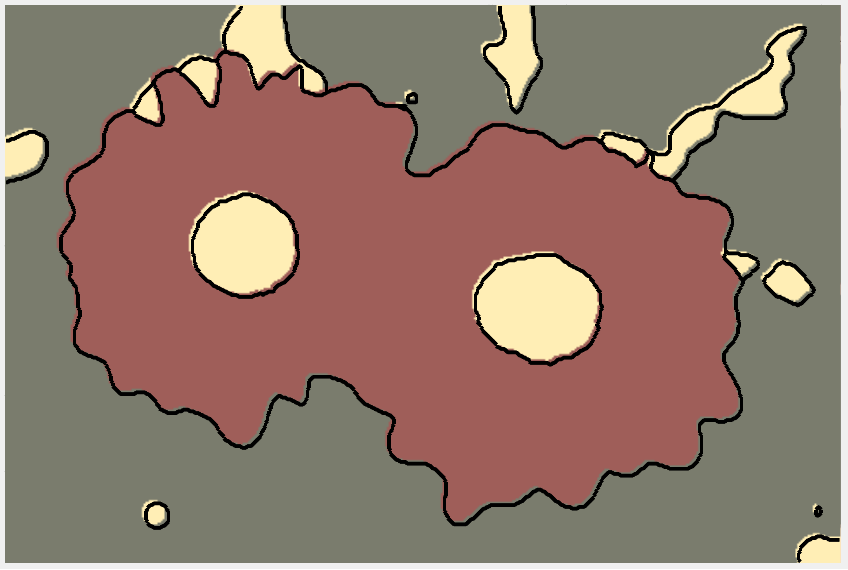} \\
\includegraphics[width=\fw]{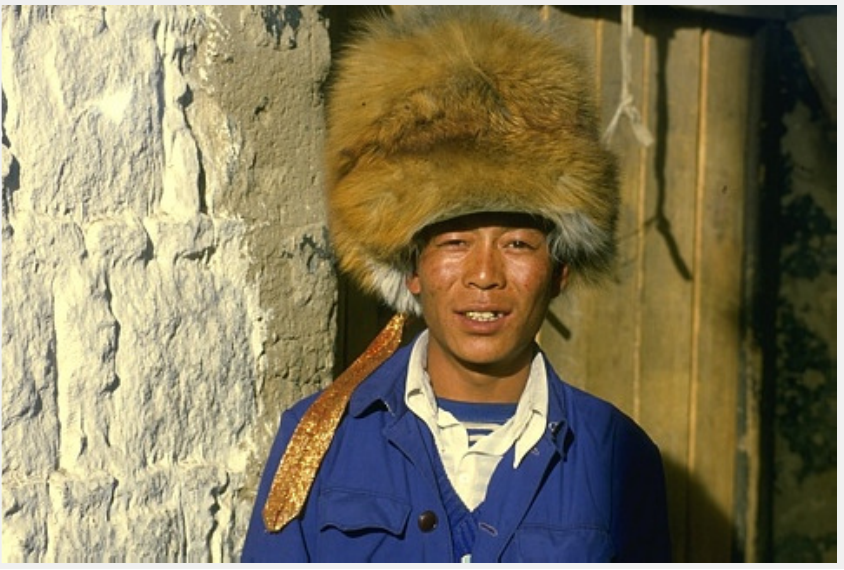} \hspace{\sp} & 
\includegraphics[width=\fw]{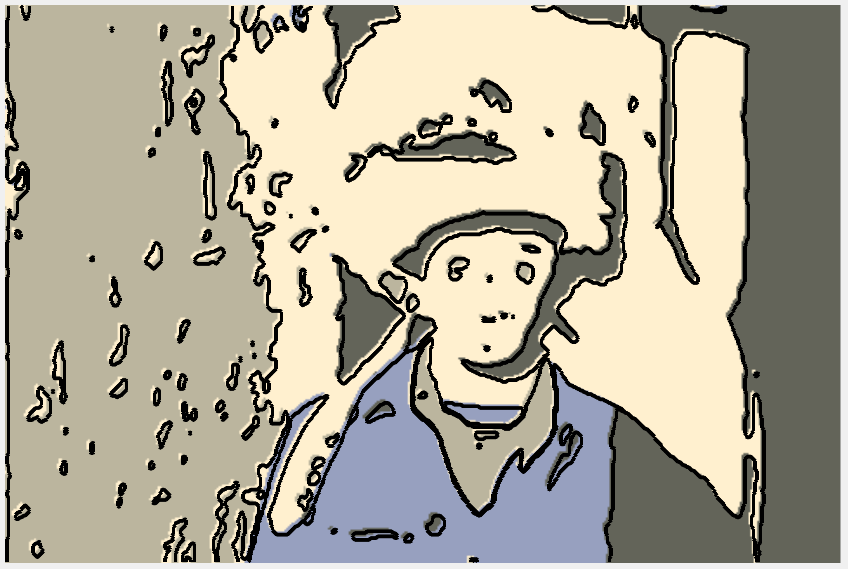} \hspace{\sp} &  
\includegraphics[width=\fw]{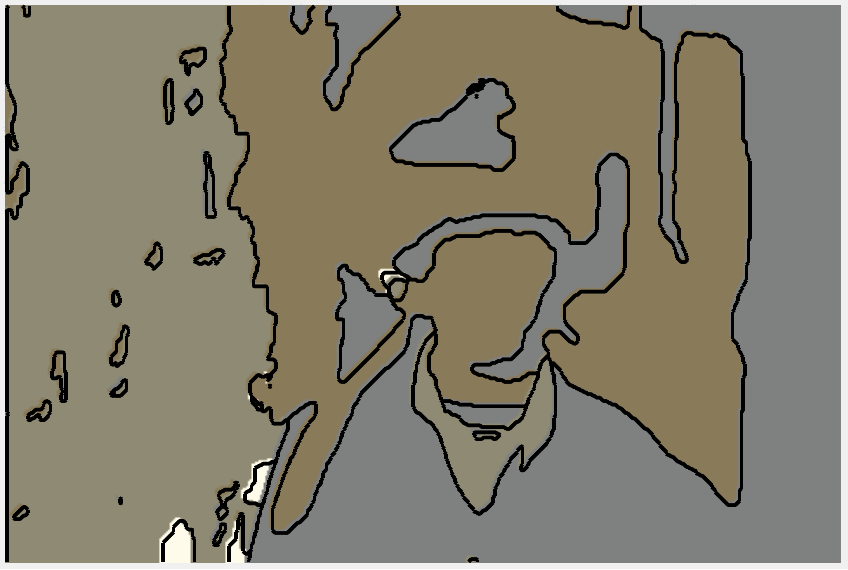} \hspace{\sp} &  
\includegraphics[width=\fw]{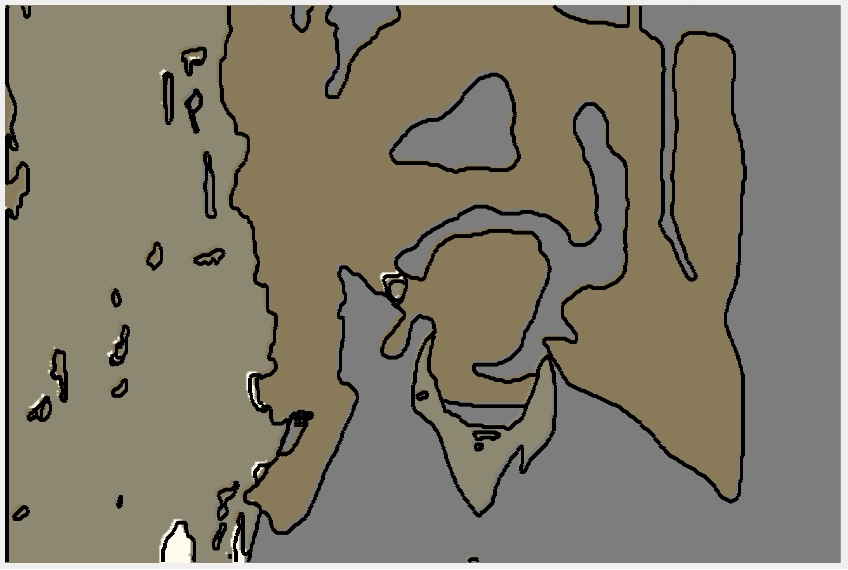} \hspace{\sp} &  
\includegraphics[width=\fw]{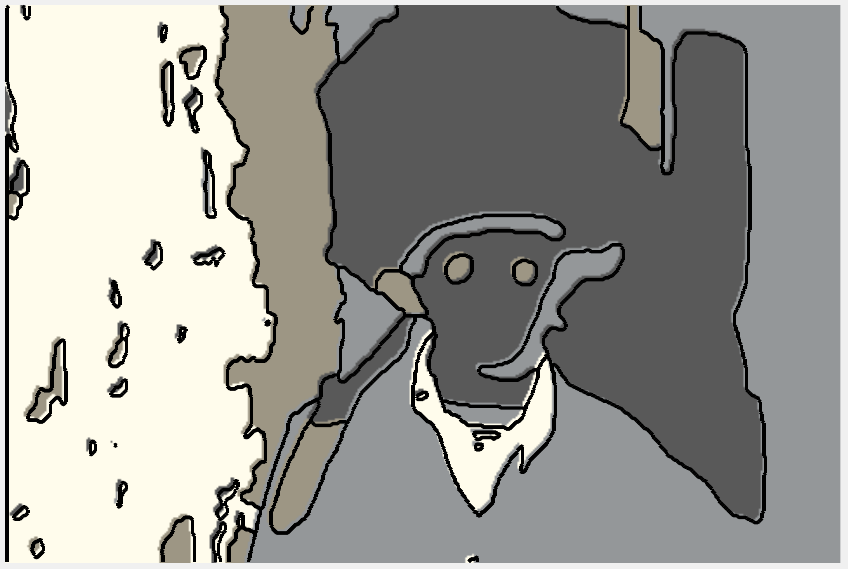} \hspace{\sp} &  
\includegraphics[width=\fw]{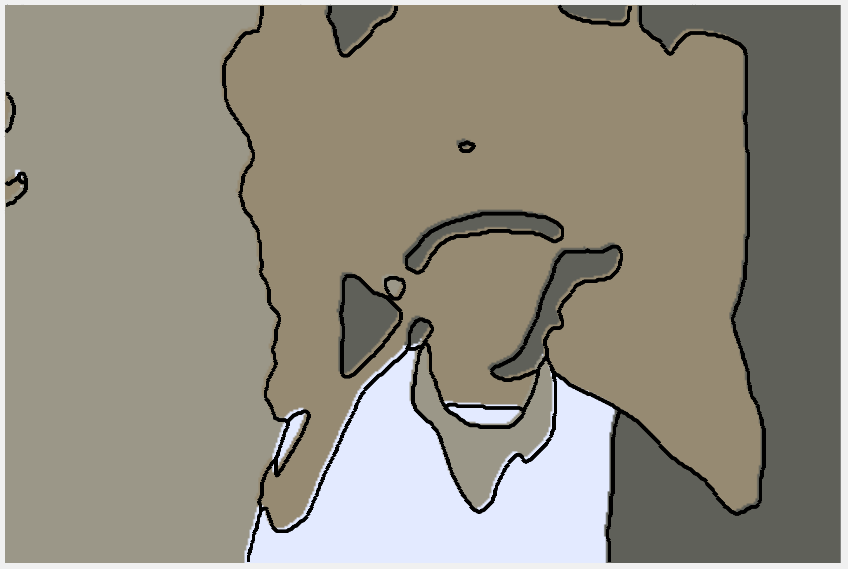} \\
\includegraphics[width=\fw]{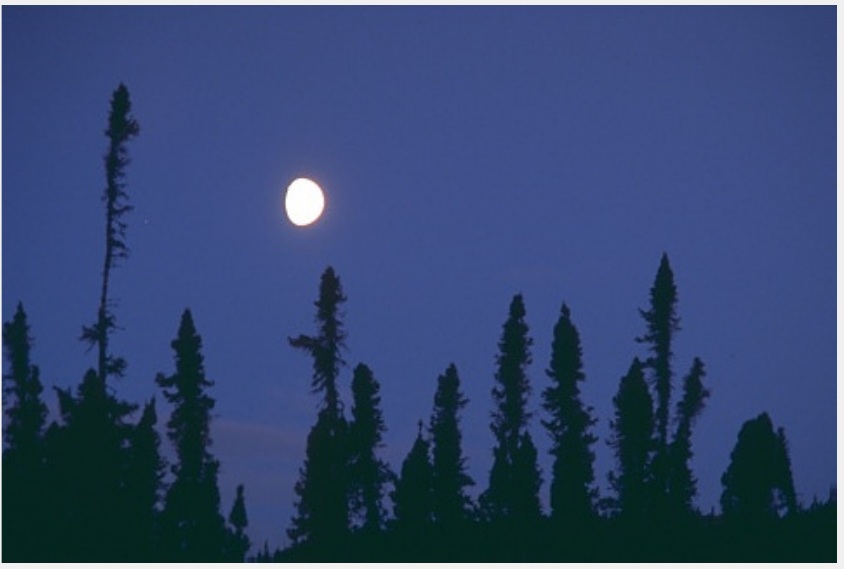} \hspace{\sp} & 
\includegraphics[width=\fw]{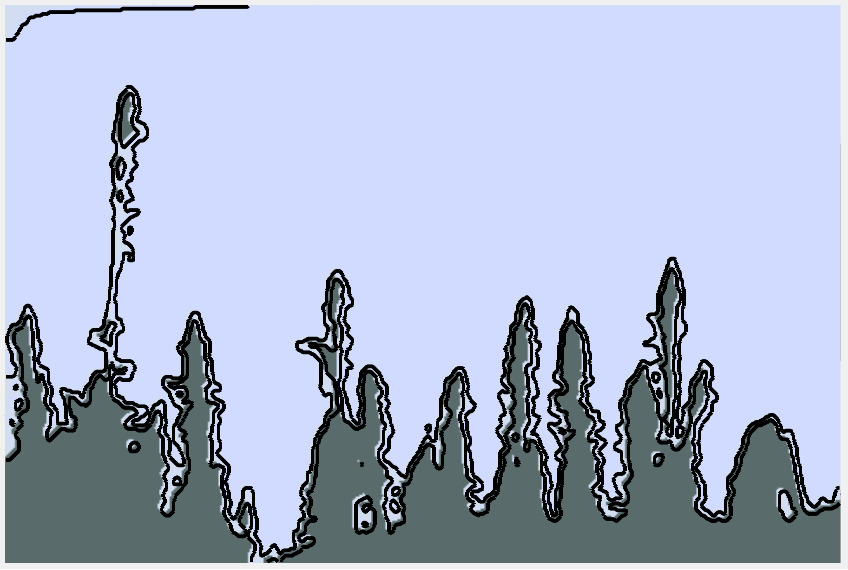} \hspace{\sp} &  
\includegraphics[width=\fw]{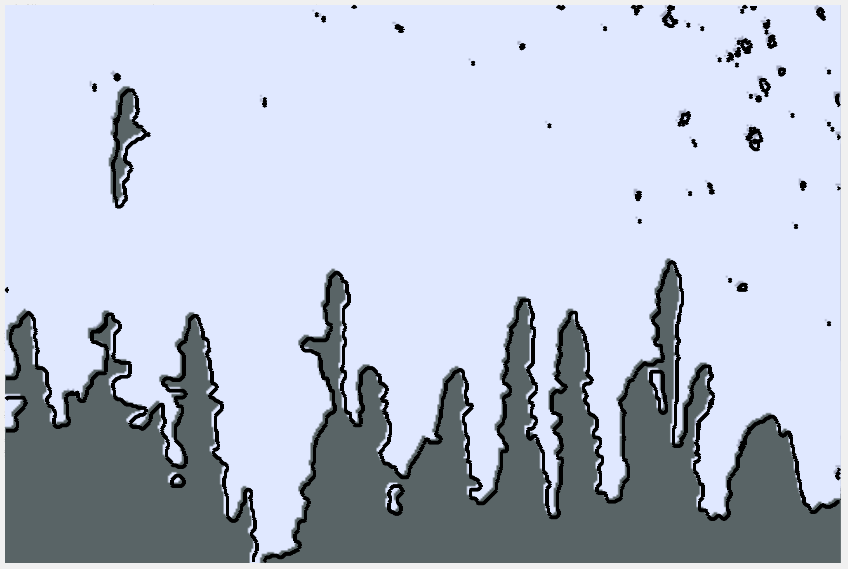} \hspace{\sp} &  
\includegraphics[width=\fw]{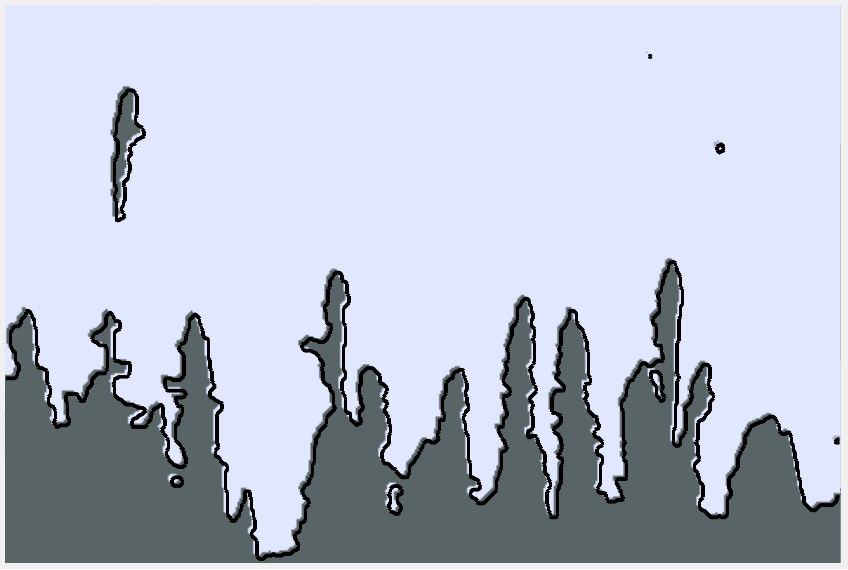} \hspace{\sp} &  
\includegraphics[width=\fw]{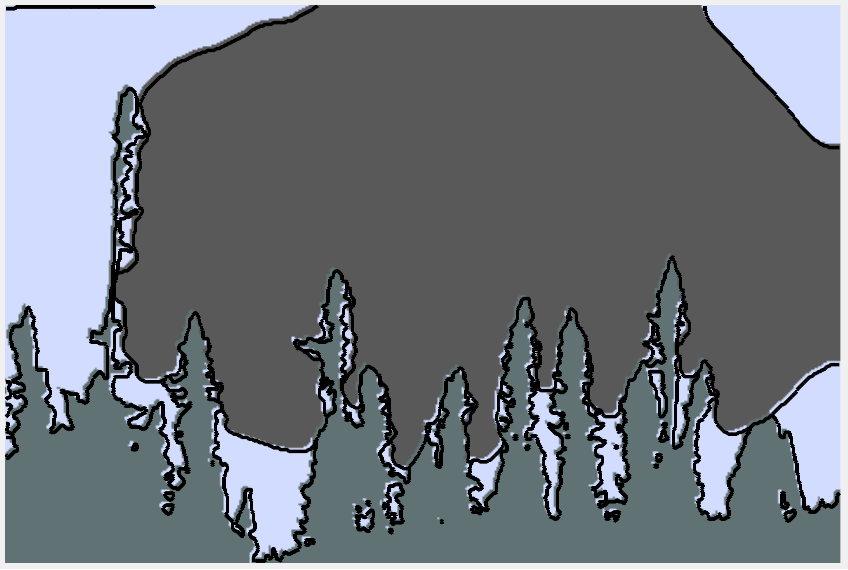} \hspace{\sp} &  
\includegraphics[width=\fw]{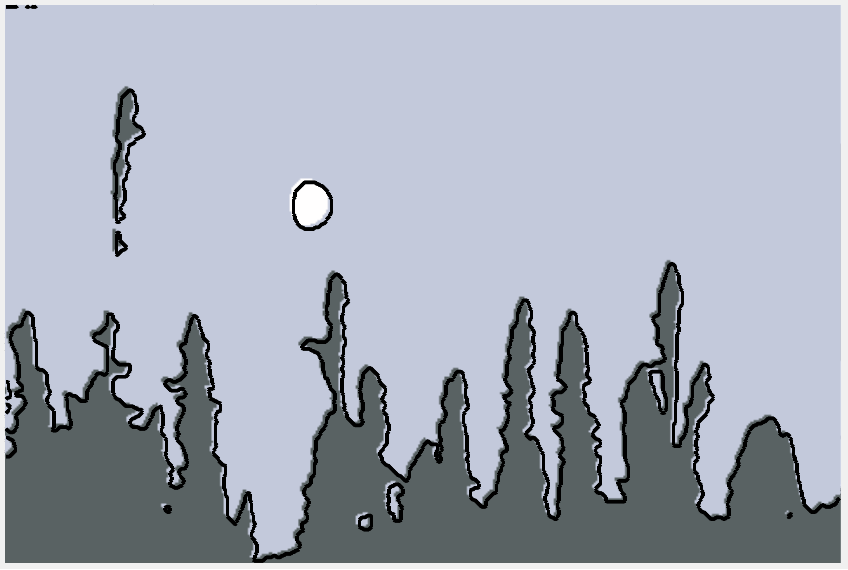} \\
\includegraphics[width=\fw]{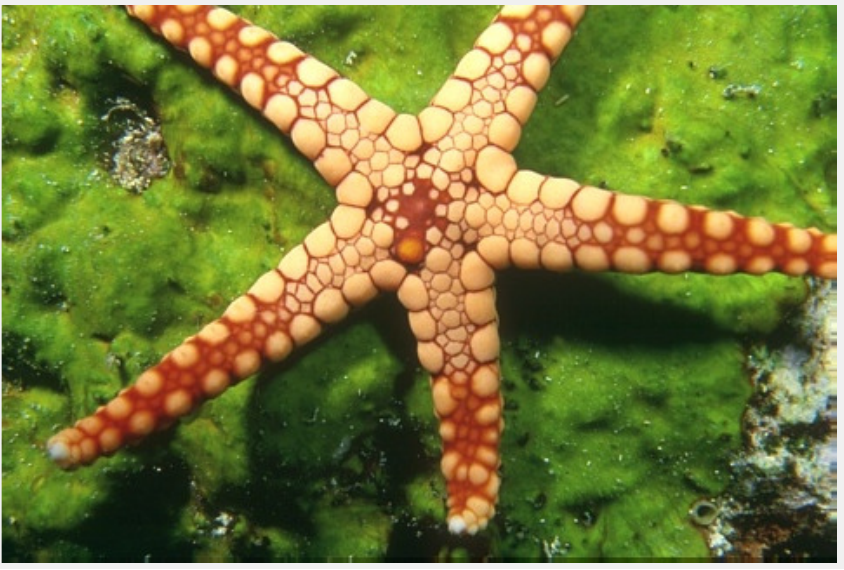} \hspace{\sp} & 
\includegraphics[width=\fw]{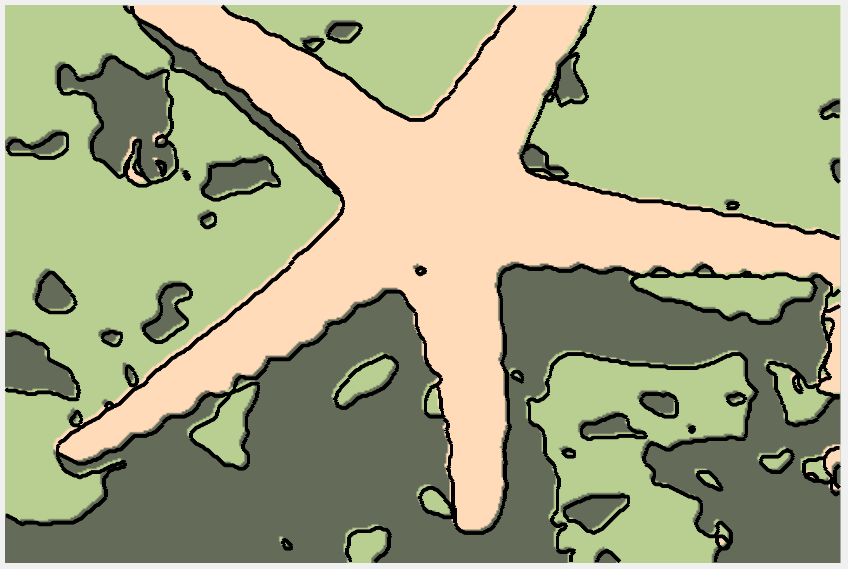} \hspace{\sp} &  
\includegraphics[width=\fw]{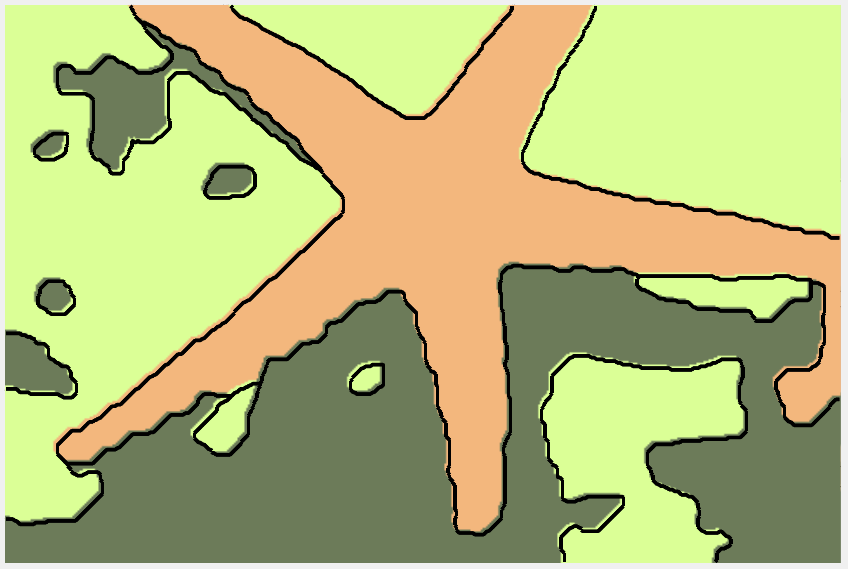} \hspace{\sp} &  
\includegraphics[width=\fw]{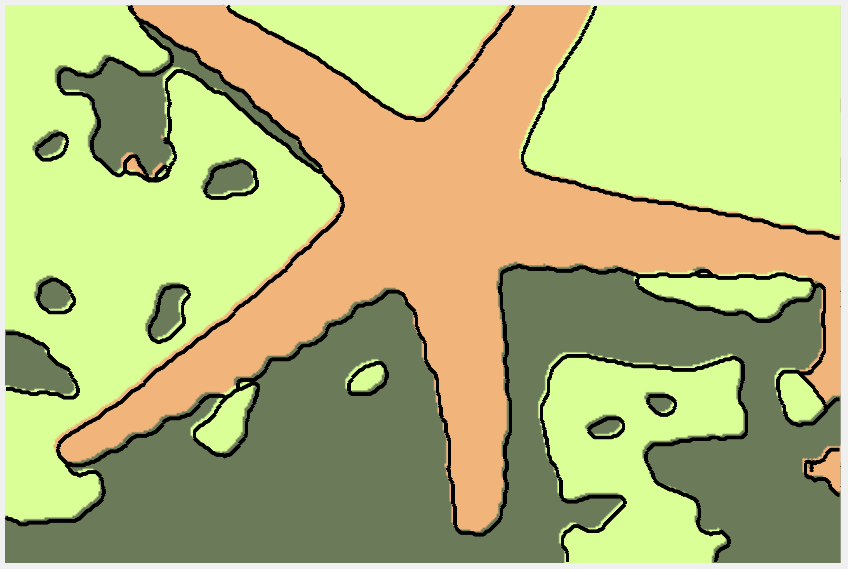} \hspace{\sp} &  
\includegraphics[width=\fw]{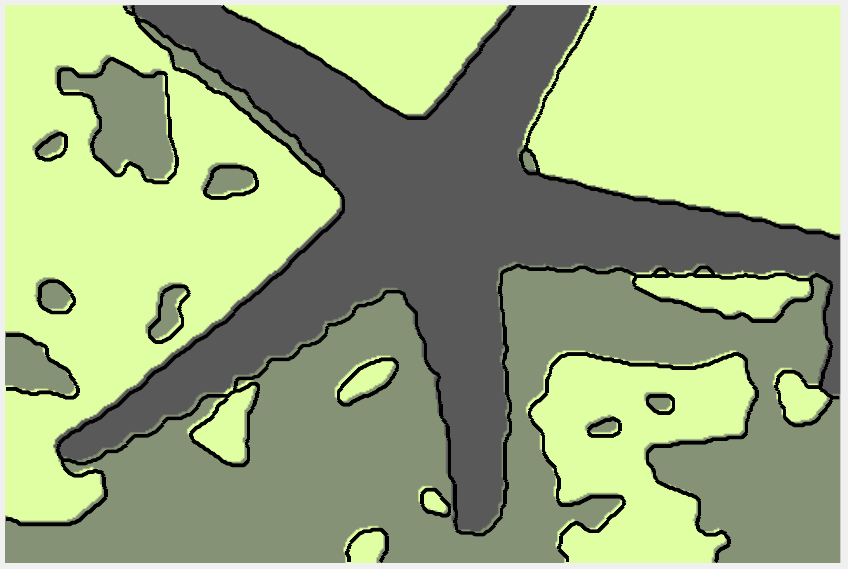} \hspace{\sp} &  
\includegraphics[width=\fw]{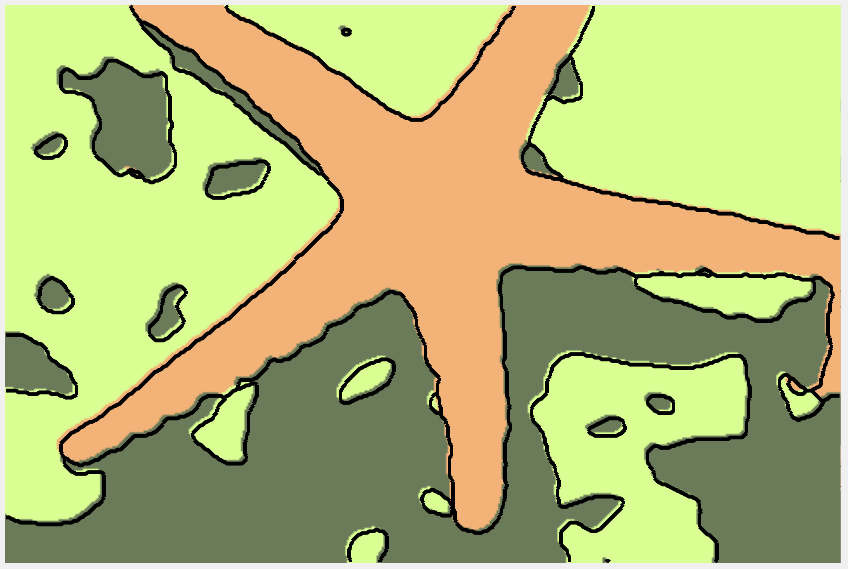} \\
\includegraphics[width=\fw]{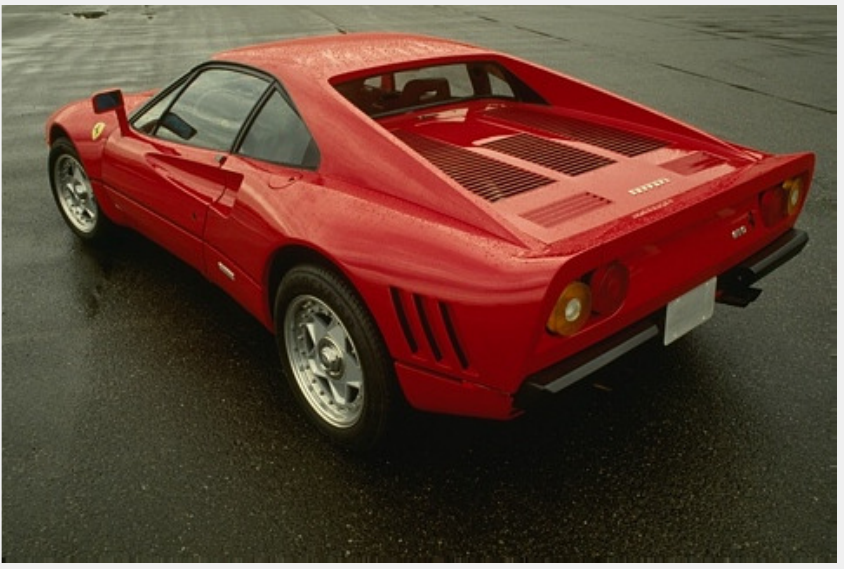} \hspace{\sp} & 
\includegraphics[width=\fw]{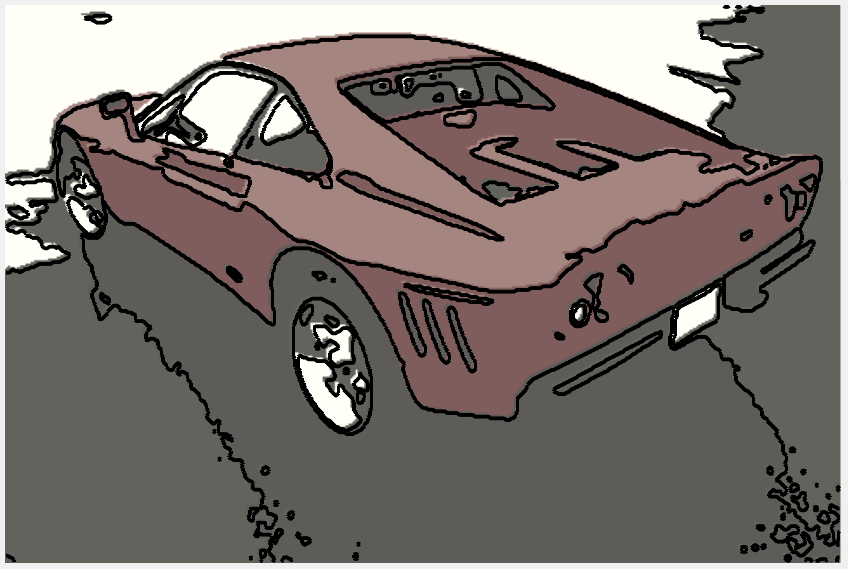} \hspace{\sp} &  
\includegraphics[width=\fw]{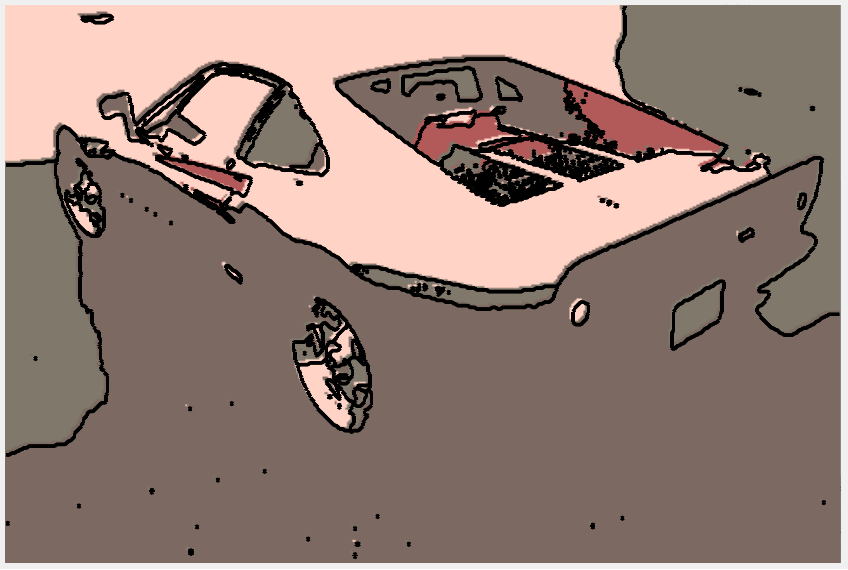} \hspace{\sp} &  
\includegraphics[width=\fw]{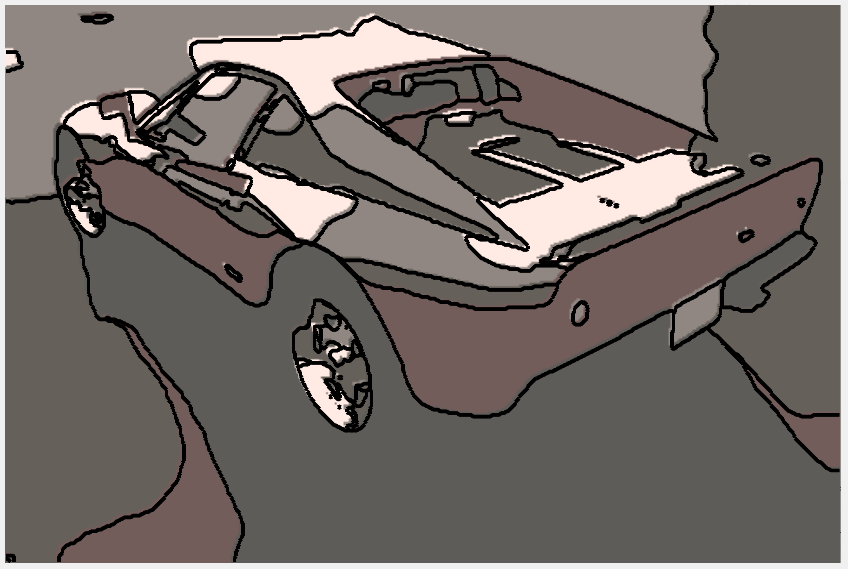} \hspace{\sp} &  
\includegraphics[width=\fw]{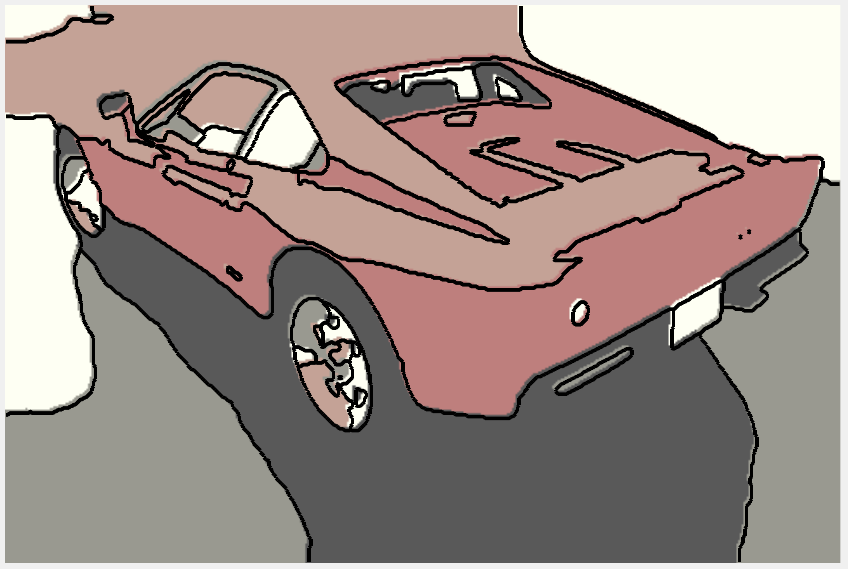} \hspace{\sp} &  
\includegraphics[width=\fw]{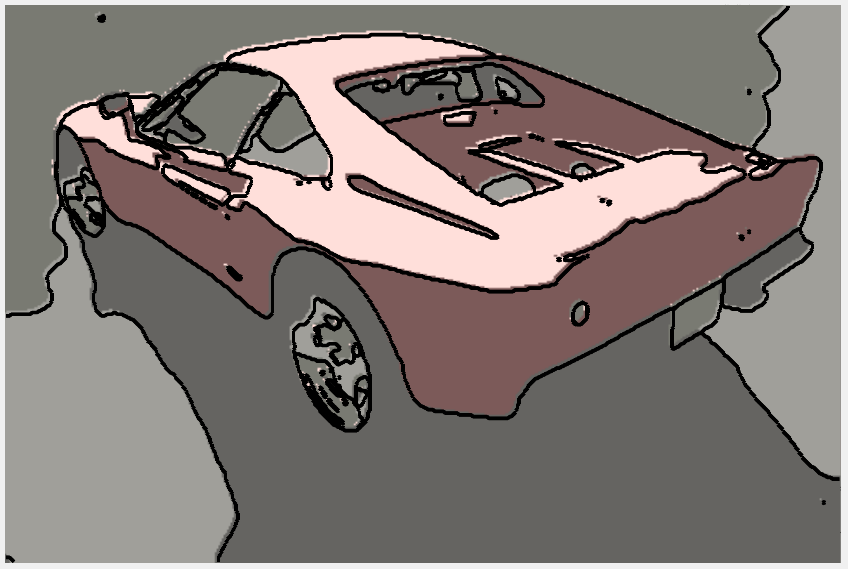} \\
(a) Input \hspace{\sp} & (b) FL~\cite{sundaramoorthi2014fast} \hspace{\sp} & (c) TV~\cite{zach2008fast} \hspace{\sp} & (d) VTV~\cite{lellmann:continuous:siam:2011} \hspace{\sp} & (e) PC~\cite{chambolle2012convex} \hspace{\sp} & (f) Ours
\end{tabular}
\caption{Visual comparison of the multi-label segmentation on Berkeley dataset using different algorithms.}
\label{fig:berkeley}
\end{figure*}
%
%
We perform a comparative analysis for the multi-label segmentation based on the Berkeley dataset~\cite{arbelaez2011contour}.
We compare our algorithm to the existing state-of-the-art techniques of which the underlying model assumes the piecewise constant image for fair comparison, and consider the algorithms: FL~\cite{sundaramoorthi2014fast}, TV~\cite{zach2008fast}, VTV~\cite{lellmann:continuous:siam:2011}, PC~\cite{chambolle2012convex}. 
%
%
%
\begin{table} 
\centering
\scriptsize
\begin{tabular}{|c|c|c|c|c|c|}
\hline
labels & FL~\cite{sundaramoorthi2014fast} & TV~\cite{zach2008fast} & VTV~\cite{lellmann:continuous:siam:2011} & PC~\cite{chambolle2012convex} & Ours\\
\hline
3&0.53$\pm$0.11&0.68$\pm$0.17&0.68$\pm$0.16&0.57$\pm$0.15&0.67$\pm$0.13\\
4&0.48$\pm$0.08&0.53$\pm$0.18&0.58$\pm$0.31&0.57$\pm$0.20&0.63$\pm$0.13\\
5&0.44$\pm$0.12&0.43$\pm$0.30&0.50$\pm$0.22&0.49$\pm$0.16&0.60$\pm$0.05\\
6&0.42$\pm$0.10&0.37$\pm$0.24&0.43$\pm$0.16&0.47$\pm$0.14&0.52$\pm$0.19\\
\hline
\end{tabular} 
\caption{Precision of the results with varying number of labels.}
\label{tab:precision}
\end{table} 
%
%
\begin{table}
\centering
\scriptsize
\begin{tabular}{|c|c|c|c|c|c|}
\hline
labels & FL~\cite{sundaramoorthi2014fast} & TV~\cite{zach2008fast} & VTV~\cite{lellmann:continuous:siam:2011} & PC~\cite{chambolle2012convex} & Ours\\
\hline
3&0.78$\pm$0.11&0.67$\pm$0.10&0.67$\pm$0.11&0.76$\pm$0.11&0.69$\pm$0.12\\
4&0.84$\pm$0.08&0.71$\pm$0.04&0.75$\pm$0.08&0.76$\pm$0.14&0.72$\pm$0.08\\
5&0.89$\pm$0.06&0.81$\pm$0.07&0.74$\pm$0.11&0.71$\pm$0.18&0.73$\pm$0.13\\
6&0.82$\pm$0.09&0.71$\pm$0.15&0.71$\pm$0.10&0.72$\pm$0.10&0.65$\pm$0.22\\
\hline
\end{tabular}
\caption{Recall of the results with varying number of labels.}
\label{tab:recall}
\end{table}
%
%
%
We provide the qualitative evaluation in Fig.~\ref{fig:berkeley} where the input images are shown in (a) and the segmentation results are shown in (b)-(f) where the same number of labels is applied to all the algorithms for each image.
The algorithm parameters for the algorithms under comparison are optimized with respect to the accuracy while we set the parameters: $\eta$=0.5, $\mu$=0.5, $\alpha$=0.01, $\beta$=10, $\tau$=0.5, $\theta$=1.
While our method yields better labels than the others, the obtained results may seem to be imperfect in general, which is due to the limitation of the underlying image model in particular in the presence of texture or illumination changes.
The quantitative comparisons are reported in terms of precision and recall with varying number of labels in Tables~\ref{tab:precision},~\ref{tab:recall}. 
%
%
The computational cost as a baseline excluding the use of special hardware (e.g. multi-core GPU/CPU) and image processing techniques (e.g. image pyramid) is provided for $481 \times 321 \times 3$ RGB images in Table~\ref{tab:cost}.
%
%
%
\begin{table}[hbt]
\centering
\scriptsize
\begin{tabular}{|c|c|c|c|c|c|c|c|c|}
\hline
\# of labels&2&3&4&5&6&7&8&9\\
\hline
time (sec)&3.08&4.40&5.82&7.16&8.62&10.07&11.66&12.72\\
\hline
\end{tabular}
\caption{Computational cost with varying number of labels.}
\label{tab:cost}
\end{table}
%
%
%
\section{Conclusion}

We have introduced a multi-label segmentation model that is motivated by Information Theory. The proposed model has been designed to make use of a prior, or regularization functional, that adapts to the residual during convergence of the algorithm in both space, and time/iteration. This results in a natural annealing schedule, with no adjustable parameters, whereby the influence of the prior is strongest at initialization, and wanes as the solution approaches a good fit with the data term. All this is done within an efficient convex optimization framework using the ADMM.
Our functional has been based on a Huber-Huber penalty, both for data fidelity and regularization. This is made efficient by a variable splitting, which has yielded faster and more accurate region boundaries in comparison to the TV-$L_1$ and TV-$L_2$ models. 
To make all this work for multi-label segmentation with a large number of labels, we had to introduce a novel constraint that penalized the common area of the pairwise region combinations. 
The proposed energy formulation and optimization algorithm can be naturally applied to a number of imaging tasks such as image reconstruction, motion estimation and motion segmentation. Indeed, segmentation can be improved by a more sophisticated data term integrated into our algorithm.
%
%

\newpage


\appendix

\section{From Bayes-Tikhonov to Information-Driven Regularization} \label{sec:justification:bayes}

This section offers an alternative motivation for our choice of adaptive regularization.

The cost function \eqref{eq:ConvexEnergy}, simplified after removing auxiliary variables used in the optimization, consists of a point-wise sum, which could be interpreted probabilistically by assuming that the image $f$ is a sample from an IID random field. 

In a Bayesian setting, one would {\em choose a prior} $q$ based on assumptions about the state of nature. For instance, assuming natural images to be piecewise smooth, one can choose $q \doteq p(u(x) | c_i) \propto \exp(- \gamma(\nabla u_i(x)))$. Once chosen a prior, one would then choose or learn a likelihood model $\ell = p(f(x) | u(x), c_i)$, or equivalently a {\em posterior} $p = \ell q$ where, for instance, $p = p(u(x) | f(x), c_i) \propto \exp(-\rho(u_i(x)))$. The natural inference criterion in a Bayesian setting would be to maximize the posterior:
\begin{equation}
\psi^{\rm max}_{\ell,q} = \ell q = p.
\end{equation}
Here the prior exerts its influence with equal strength throughout the inference process, and is generally related to standard Tikhonov regularization. Instead, we would like a natural annealing process that starts with the prior as our initial model, and gradually increases the influence of the data term until, at convergence, the influence of the prior is minimized and the data drives convergence to the final solution. This could be done by an annealing parameter $\lambda$ that changes from $0$ to $1$ according to some schedule, yielding (an homotopy class of cost functions mapping the prior to the likelihood as $\lambda$ goes from $0$ to $1$)
\begin{equation}
\psi^{\rm max}_{\ell,q}(\lambda) = \ell^\lambda q^{(1-\lambda)} \neq p
\label{eq:noBayes}
\end{equation}
Ideally, we would like to not pick this parameter by hand, but instead have the data guide the influence of the prior during the optimization. Bayesian inference here not only does not provide guidance on how to choose the annealing schedule, but it does not even allow a changing schedule, for that would not be compatible with a maximum a-posteriori inference criterion. 

Therefore, we adopt an information-theoretic approach: We first choose the prior $q$, just like in the Bayesian case, but then we seek for a likelihood $\ell$, or equivalently a posterior $p$, {\em that makes the data as informative as possible.} This is accomplished by choosing the posterior $p$ to be as divergent as possible from the prior, in the sense of Kullback-Leibler. That way, the data is as informative as possible (if the posterior was chosen to be the same as the prior, the system would not even look at the data).  This yields a natural cost criterion
\begin{equation}
\phi^{\rm max}_{p,q} = {\mathbb{KL}}(p || q) + {\rm const.}
\end{equation}
choosing the constant to be ${\mathbb E}(\log q)$, we obtain the equivalent cost function to be minimized:
\begin{equation}
\phi^{\rm min}_{p,q} = - {\mathbb E}\left(p \log \frac{p}{q}\right) + {\mathbb E}(\log q)
\end{equation}
where the expectation is with respect to the variability of the data sampled on the spatial domain. If such samples are assumed IID, we have that 
\begin{equation}
\phi^{\rm min}_{p,q} = - \log \psi_{\ell, q}^{\rm max}(p)
\end{equation}
from \eqref{eq:noBayes}. Thus the information-theoretic approach suggests a scheduling of the annealing parameter $\lambda$ that is data-driven, and proportional to the posterior $p$.

This yields a model with adaptive regularization that automatically adjusts during the optimization: The influence of the prior $q$ wanes as the solution becomes an increasingly better fit of the data, without the undesirable bias of the prior on the final solution (see Fig. \ref{fig:compare:model:constraint}). At the same time, it benefits from heavy regularization at the outset, turning standard regularization into a form of relaxation, annealing, or homotopy continuation method. 

In practice, the most tangible advantages of this choice are an algorithm that is easy to tune, and never again having to answer the question: {\em ``how did you choose $\lambda$?''}

%
%
\section{Detailed Derivation of Energy Optimization} \label{sec:optimization:detail}

The energy~\eqref{eq:ConvexEnergyAdaptive} is minimized with respect to the partitioning functions $u_i$ and the intensity estimates $c_i$ in an alternating way. Since it is convex in $u_i$ with fixed $c_i$, we use an efficient convex optimization algorithm in the framework of alternating-direction method of multipliers (ADMM)~\cite{boyd2011distributed,parikh2014proximal}.
We modify the energy functional in~\eqref{eq:ConvexEnergyAdaptive} using variable splitting~\cite{courant1943variational,eckstein1992douglas,wang2008new} introducing a new variable $v_i$ with the constraint $u_i = v_i$ as follows:
\begin{align} \label{eq:ConvexEnergySplit:appen}
&\sum_{i \in \Lambda} \bigg\{ \int_\Omega \lambda_i \, \rho(u_i; c_i) + \tau \Big( \sum_{i \neq j} u_j \Big) u_i \ud x \nonumber\\
&+ \int_\Omega (1 - \lambda_i) \, \gamma( \nabla v_i ) \ud x + \frac{\theta}{2} \| u_i - v_i + y_i \|_2^2 \bigg\}, \nonumber\\
&\textrm{subject to } u_i(x) \ge 0, \, \sum_{i \in \Lambda} v_i(x) = 1, \, \forall x \in \Omega,
\end{align}
where $y_i$ is a dual variable for each equality constraint $u_i = v_i$, and $\theta > 0$ is a scalar augmentation parameter.
The original constraints $u_i \in [0, 1]$ and $\sum_i u_i = 1$ in~\eqref{eq:ConvexEnergyAdaptive} are decomposed into the simpler constraints $u_i \ge 0$ and $\sum_i v_i = 1$ in~\eqref{eq:ConvexEnergySplit} by variable splitting $u_i = v_i$.
In the computation of the data fidelity and the regularization, we employ a robust estimator using the Huber loss function defined in~\eqref{eq:HuberFunction}. 
An efficient procedure can be performed to minimize the Huber loss function $\phi_{\eta}$ following the equivalence property of Moreau-Yosida regularization of a non-smooth function $| \cdot |$ as defined by~\cite{moreau1965proximite,opac-b1133911}:
\begin{align} \label{eq:MoreauYosida:appen}
\phi_{\eta}(x) = \inf_r \left\{ | r | + \frac{1}{2 \eta} (x - r)^2 \right\},
\end{align}
which replaces the data fidelity $\rho(u_i; c_i)$ and the regularization $\gamma(\nabla v_i)$ in~\eqref{eq:ConvexEnergySplit} with the regularized forms $\rho(u_i; c_i, r_i)$ and $\gamma(\nabla v_i; z_i)$, respectively as follows:
\begin{align} 
\rho(u_i; c_i, r_i) &= \left( | r_i | + \frac{1}{2 \eta} (f - c_i - r_i)^2 \right) u_i, \label{eq:DataFidelityMoreauYosida:appen}\\
\gamma(\nabla v_i; z_i) &=  \| z_i \|_1 + \frac{1}{2 \mu} \| \nabla v_i - z_i \|^2_2, \label{eq:RegularizationMoreauYosida:appen}
\end{align}
where $r_i$ and $z_i$ are the auxiliary variables to be minimized in alternation.
The constraints on $u_i$ and $v_i$ in~\eqref{eq:ConvexEnergySplit} can be represented by the indicator function $\delta_A(x)$ of a set $A$ defined by:
\begin{align} \label{eq:DeltaFunction:appen}
\delta_A(x) &= 
   \begin{cases}
   0 & : x \in A\\
   \infty & : x \notin A\\
   \end{cases}
\end{align}
The constraint $u_i \ge 0$ is given by $\delta_A(u_i)$ where $A = \{ x | x \ge 0 \}$, and the constraint $\sum_i v_i = 1$ is given by $\delta_B( \{ v_i \} )$ where $B = \left\{ \{x_i\} | \sum_i x_i = 1 \right\}$.
The Moreau-Yosida regularization for the Huber loss function and the constraints represented by the indicator functions lead to the following unconstrained energy functional $\mathcal{L}_i$ for label $i$:
\begin{align} \label{eq:EnergyUnconstrainedLabel:appen}
&\mathcal{L}_i = \int_\Omega \lambda_i \, \rho(u_i; c_i, r_i) + \tau \Big( \sum_{i \neq j} u_j \Big) u_i \ud x \nonumber + \delta_A(u_i)\\
&+ \int_\Omega (1 - \lambda_i) \, \gamma( \nabla v_i; z_i ) \ud x + \frac{\theta}{2} \| u_i - v_i + y_i \|_2^2,
\end{align}
and the final unconstrained energy functional $\mathcal{E}$ reads:
\begin{align} \label{eq:EnergyUnconstrained:appen}
\mathcal{E}(\{u_i, v_i, y_i, c_i, r_i, z_i\}) = \sum_{i \in \Lambda} \mathcal{L}_i + \delta_B(\{v_i\}).
\end{align}
The optimal set of partitioning functions $\{ u_i \}$ is obtained by minimizing the energy functional $\mathcal{E}$ using ADMM, minimizing the augmented Lagrangian $\mathcal{L}_i$ in~\eqref{eq:EnergyUnconstrainedLabel} with respect to the variables $u_i, v_i, c_i, r_i, z_i$, and applying a gradient ascent scheme with respect to the dual variable $y_i$ followed by the update of the weighting function $\lambda_i$ and the projection of $\{ v_i \}$ onto the set $B$ in~\eqref{eq:EnergyUnconstrained}. 
The alternating optimization steps using ADMM are presented in Algorithm~\ref{alg:admm}, where $k$ is the iteration counter.
%
%
\begin{algorithm}[tb]
\caption{The ADMM updates for minimizing~\eqref{eq:EnergyUnconstrained}}
\label{alg:admm:appen}
\begin{algorithmic}
\State 
\textbf{for} each label $i \in \Lambda$ \textbf{ do}
\begin{flalign}
\;\; c_i^{k+1} & \coloneqq \argmin_{c} \rho( u_i^k; c, r_i^k ) & \label{step:c:appen}\\
\;\; r_i^{k+1} & \coloneqq \argmin_{r} \rho( u_i^k; c_i^{k+1}, r ) & \label{step:r:appen}\\
\;\; \nu_i^{k+1} & \coloneqq \exp\left( - \frac{\rho(u_i^k; c_i^{k+1}, r_i^{k+1})}{\beta} \right) & \label{step:nu:appen}\\
\;\; \lambda_i^{k+1} & \coloneqq \argmin_{\lambda} \frac{1}{2} \| \nu_i^{k+1} - \lambda \|_2^2 + \alpha \| \lambda \|_1 & \label{step:l:appen}\\
\;\; z_i^{k+1} & \coloneqq \argmin_z \gamma(\nabla v_i^k; z) & \label{step:z:appen}\\
\;\; u_i^{k+1} & \coloneqq \argmin_{u} \int_\Omega \lambda_i^{k+1} \rho(u; c_i^{k+1}, r_i^{k+1}) \ud x + \delta_A(u) & \nonumber\\
& \hspace{-6pt} + \int_\Omega \tau \Big( \sum_{i \neq j} u_j \Big) u \ud x + \frac{\theta}{2} \| u - v_i^k + y_i^k \|_2^2  & \label{step:u:appen}\\
\;\; \tilde{v}_i^{k+1} & \coloneqq \argmin_{v} \int_\Omega \left( 1-\lambda_i^{k+1} \right) \gamma(\nabla v; z_i^{k+1}) \ud x & \nonumber\\
& \hspace{-6pt} + \frac{\theta}{2} \| u_i^{k+1} - v + y_i^k \|_2^2 & \label{step:v:appen}\\
y_i^{k+1} & \coloneqq y_i^k + u_i^{k+1} - v_i^{k+1} & \label{step:y:appen}
\end{flalign}
\textbf{end for}
\begin{flalign}
\{ v_i^{k+1} \} & \coloneqq \Pi_B \left( \{ \tilde{v}_i^{k+1} \} \right) & \label{step:v:projection:appen}
\end{flalign}
\end{algorithmic}
\end{algorithm}
%
%
The update of the estimate $c_i^{k+1}$ in~\eqref{step:c} is obtained by the following step:
\begin{align} 
\quad c_i^{k+1} \coloneqq \frac{\int_\Omega \lambda_i^k (f - r_i^k) u_i^k \ud x}{\int_\Omega \lambda_i^k u_i^k \ud x}. \label{step:SolutionC:appen}
\end{align}
The update for the auxiliary variable $r_i^{k+1}$ in~\eqref{step:r} is obtained by the following optimality condition:
\begin{align} 
0 & \in \partial | r_i^{k+1} | - \frac{1}{\eta} ( f - c_i^{k+1} - r_i^{k+1} ), \label{eq:OptimalityR:appen}
\end{align}
where $\partial$ denotes the sub-differential operator. 
The solution for the optimality condition in~\eqref{eq:OptimalityR} is obtained by the proximal operator of the $L_1$ norm~\cite{parikh2014proximal} as follows:
\begin{align}
r_i^{k+1} \coloneqq \prox \left( f - c_i^{k+1} \, \big| \, \eta \, g \right), \label{eq:prox_r:appen}
\end{align}
where $g(x) = \| x \|_1$. 
The proximal operator $\prox(v \, | \, \eta \, g)$ of the weighed $L_1$ norm $\eta \, g$ with a parameter $\eta > 0$ at $v$ is defined by:
\begin{align}
\prox(v \, | \, \eta \, g) \coloneqq \argmin_x \left( \frac{1}{2} \| x - v \|_2^2 + \eta \, g(x) \right). \label{eq:prox:appen}
\end{align}
The solution of the proximal operator of the $L_1$ norm is obtained by the soft shrinkage operator $\mathcal{T} (v \, | \, \eta)$ defined in~\eqref{eq:shrink}.
Thus, the solutions of the proximal problem in~\eqref{eq:prox_r:appen} is given by:
\begin{align}
r_i^{k+1} \coloneqq \mathcal{T} \left( f - c_i^{k+1} \, \big| \, \eta \right). \label{eq:shrink_r:appen}
\end{align}
In the same way, the update of the auxiliary variable $z_i^{k+1}$ in~\eqref{step:z} is obtained by:
\begin{align}
z_i^{k+1} \coloneqq \mathcal{T} \left( \nabla v_i^k \, \big| \, \mu \right). \label{eq:shrink_z:appen}
\end{align}
For the primal variable $u_i^{k+1}$ in~\eqref{step:u}, we employ the intermediate solution $\tilde{u}_i^{k+1}$ and its optimality condition is given by:
\begin{align} 
0 \in \lambda_i^{k+1} d_i^{k+1} + \tau \sum_{i \neq j} u_j^k + \theta (\tilde{u}_i^{k+1} - v_i^k + y_i^k), \label{eq:StepUCondition:appen}\\
d_i^{k+1} \coloneqq | r_i^{k+1} | + \frac{1}{2 \eta} (f - c_i^{k+1} - r_i^{k+1})^2, \label{eq:StepUConditionD:appen}
\end{align}
leading to the following update:
\begin{align} 
\tilde{u}_i^{k+1} \coloneqq v_i^k - y_i^k - \frac{\lambda_i^{k+1}}{\theta} d_i^{k+1} - \frac{\tau}{\theta} \sum_{i \neq j} u_j^k. \label{eq:StepUInter:appen}
\end{align}
Given the intermediate solution $\tilde{u}_i^{k+1}$, the positivity constraint is imposed to obtain the solution for the update of $u_i^{k+1}$ as follows:
\begin{align} 
u_i^{k+1} \coloneqq \Pi_A( \tilde{u}_i^{k+1} ) = \max \{ 0, \tilde{u}_i^{k+1} \}, \label{eq:StepU:appen}
\end{align}
where the orthogonal projection operator $\Pi_A$ on a set $A = \{ x \, | \, x \ge 0 \}$ is defined by:
\begin{align} \label{eq:projection:appen}
\Pi_A(x) = \arg\min_{y \in A} \| y - x \|_2.
\end{align}
We also employ the intermediate solution $\tilde{v}_i^{k+1}$ in~\eqref{step:v} for the update of the primal variable $v_i^{k+1}$ in~\eqref{step:v:projection}. 
The optimality condition for the update of $\tilde{v}_i^{k+1}$ reads:
\begin{align} 
0 \in \frac{1 - \lambda_i^{k+1}}{\mu} \nabla^* ( \nabla \tilde{v}_i^{k+1} - z_i^{k+1} ) - \theta( u_i^{k+1} - \tilde{v}_i^{k+1} + y_i^k ), \nonumber
\end{align}
where $\nabla^*$ denotes the adjoint operator of $\nabla$, leading to the following linear system of equation with $\xi_i = \frac{1 - \lambda_i^{k+1}}{\mu \theta}$:
\begin{align} 
\tilde{v}_i^{k+1} -  \xi_i \Delta \tilde{v}_i^{k+1} = u_i^{k+1} + y_i^{k} - \xi_i \dv{z_i^{k+1}}, \label{eq:StepV:appen}
\end{align}
where $- \nabla^* \nabla = \Delta$ is the Laplacian operator, and $- \nabla^* = \mbox{div}$ is the divergence operator. 
We use the Gauss-Seidel iterations to solve the linear system in~\eqref{eq:StepV}. 
Given the set of intermediate solution $\{ \tilde{v}_i^{k+1} \}$, the solution for the update of the variable $v_i^{k+1}$ is obtained by the orthogonal projection of the intermediate solution to the set $B$ in Eq.~\eqref{step:v:projection}:
\begin{align} 
v_i^{k+1} = \tilde{v}_i^{k+1} - \frac{1}{n} \left( V - 1 \right), \quad
V = \sum_{i \in \Lambda} \tilde{v}_i^{k+1}.
\end{align}
The gradient ascent scheme is applied to update the dual variable $y_i^{k+1}$ in~\eqref{step:y}.


\newpage
{\small
\bibliographystyle{ieee}
\bibliography{multilabel}
}

\end{document}